\documentclass{article}

\usepackage[final]{neurips_data_2024}

\usepackage[pdfencoding=auto]{hyperref}

\usepackage{amsmath}
\usepackage[utf8]{inputenc} %
\usepackage[T1]{fontenc}    %
\usepackage{hyperref}       %
\usepackage{url}            %
\usepackage{booktabs}       %
\usepackage{amsfonts}       %
\usepackage{nicefrac}       %
\usepackage{microtype}      %
\usepackage{xcolor}         %
\usepackage{graphicx}
\usepackage{tikz}

\usepackage{amsthm}
\usepackage{amssymb}
\usepackage{enumitem}
\usepackage{txfonts}
\usepackage{float}
\usepackage{wrapfig}
\usepackage[breakable]{tcolorbox}
\tcbuselibrary{listingsutf8}
\usepackage{multicol}
\usepackage{tabularx}
\usepackage{tabulary}
\usepackage{titlesec}
\usepackage{svg}
\usepackage{amssymb}%
\usepackage{pifont}%
\newcommand{\cmark}{\ding{51}}%
\newcommand{\xmark}{\ding{55}}%
\titlespacing*{\section}{0pt}{0.1\baselineskip}{0.12\baselineskip}

\usetikzlibrary{positioning,arrows}
 \usepackage{multirow}

\definecolor{purple}{RGB}{230,208,222}
\definecolor{ivory}{RGB}{255,252,235}
\definecolor{nicered}{RGB}{205,86,82}
\definecolor{nicegreen}{RGB}{55,110,0}
\definecolor{niceblue}{RGB}{28, 55, 114}
\definecolor{green}{RGB}{213,232,212}
\definecolor{nicecol}{RGB}{57, 97, 107}
\definecolor{nicecol1}{RGB}{72, 74, 137}
\definecolor{nicecol2}{RGB}{119, 73, 138}
\definecolor{nicecol3}{RGB}{62, 106, 56}
\definecolor{nicecol4}{RGB}{107, 82, 57}

\theoremstyle{definition}
\newtheorem{definition}{Definition}

\newcommand\ambrosia{\raisebox{-4pt}[0cm]{\includegraphics[width=1.25em]{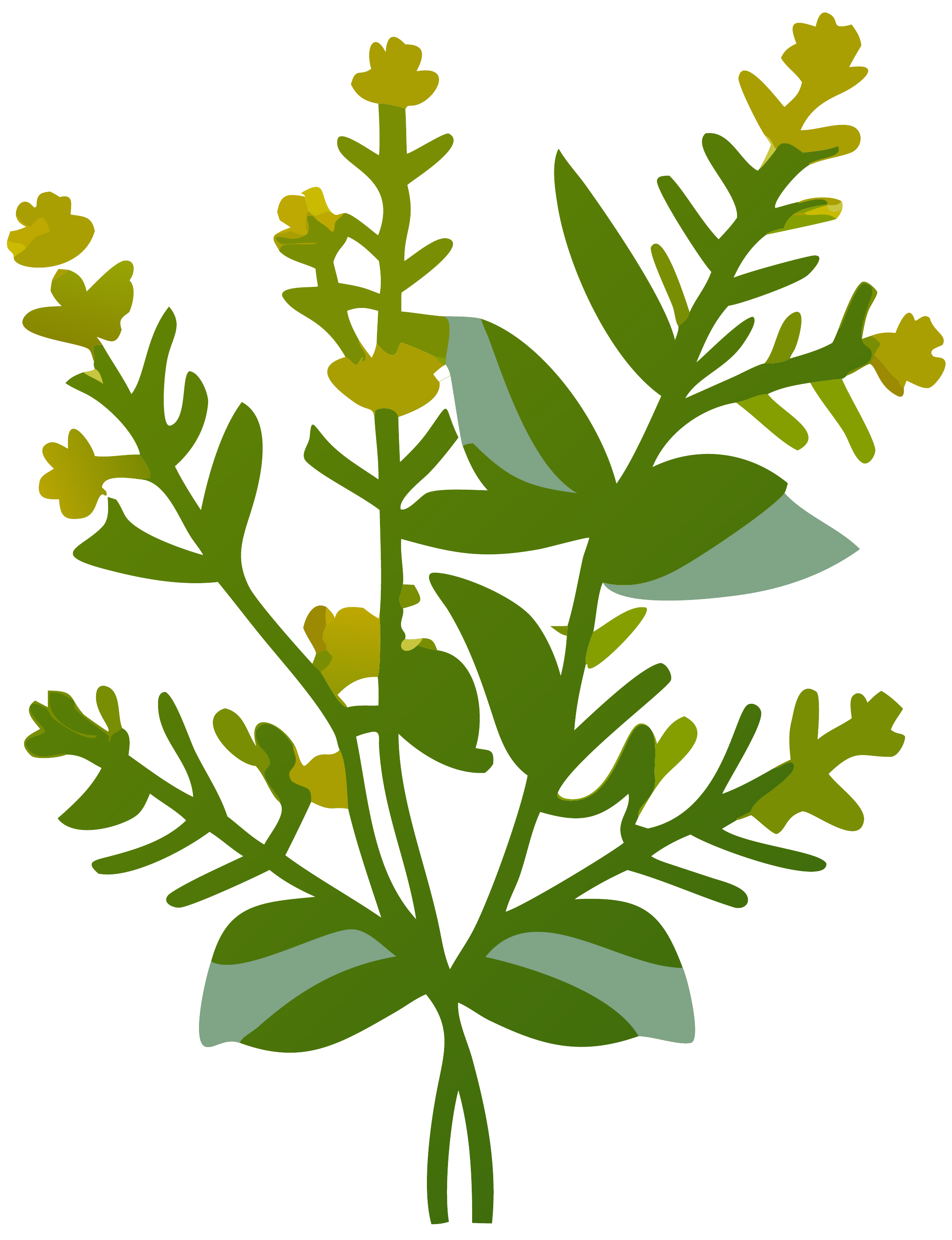}}}

\newcommand\dataset{$\varmathbb{AMBROSIA}$}

\title{\ambrosia{}\hspace{4pt}$\varmathbb{AMBROSIA}$:  A Benchmark for Parsing Ambiguous  Questions into Database Queries}

\author{
  Irina Saparina \qquad Mirella Lapata \\
  Institute for Language, Cognition and Computation \\
  School of Informatics, University of Edinburgh \\
  10 Crichton Street, Edinburgh EH8 9AB \\
  \texttt{i.saparina@sms.ed.ac.uk} \qquad \texttt{mlap@inf.ed.ac.uk}
}

\begin{document}

\maketitle

\begin{abstract}
 Practical semantic parsers are expected to understand user utterances and map them to
executable programs, even when these are ambiguous.  We introduce
  a new benchmark, \dataset{},
  which we hope will inform and inspire the development of text-to-SQL parsers capable of recognizing and interpreting ambiguous requests. Our dataset contains questions showcasing three different types of ambiguity (scope ambiguity, attachment ambiguity, and vagueness), their interpretations, and corresponding SQL queries. 
  In each case, the ambiguity persists even when the
  database context is provided. This is achieved through a novel
  approach that involves controlled generation of databases from
  scratch.  We benchmark various LLMs on \dataset{}, revealing
  that even the most advanced models struggle to identify
  and interpret ambiguity in questions. 
\end{abstract}

\section{Introduction}
\begin{figure}[t]
\includegraphics[width=\textwidth]{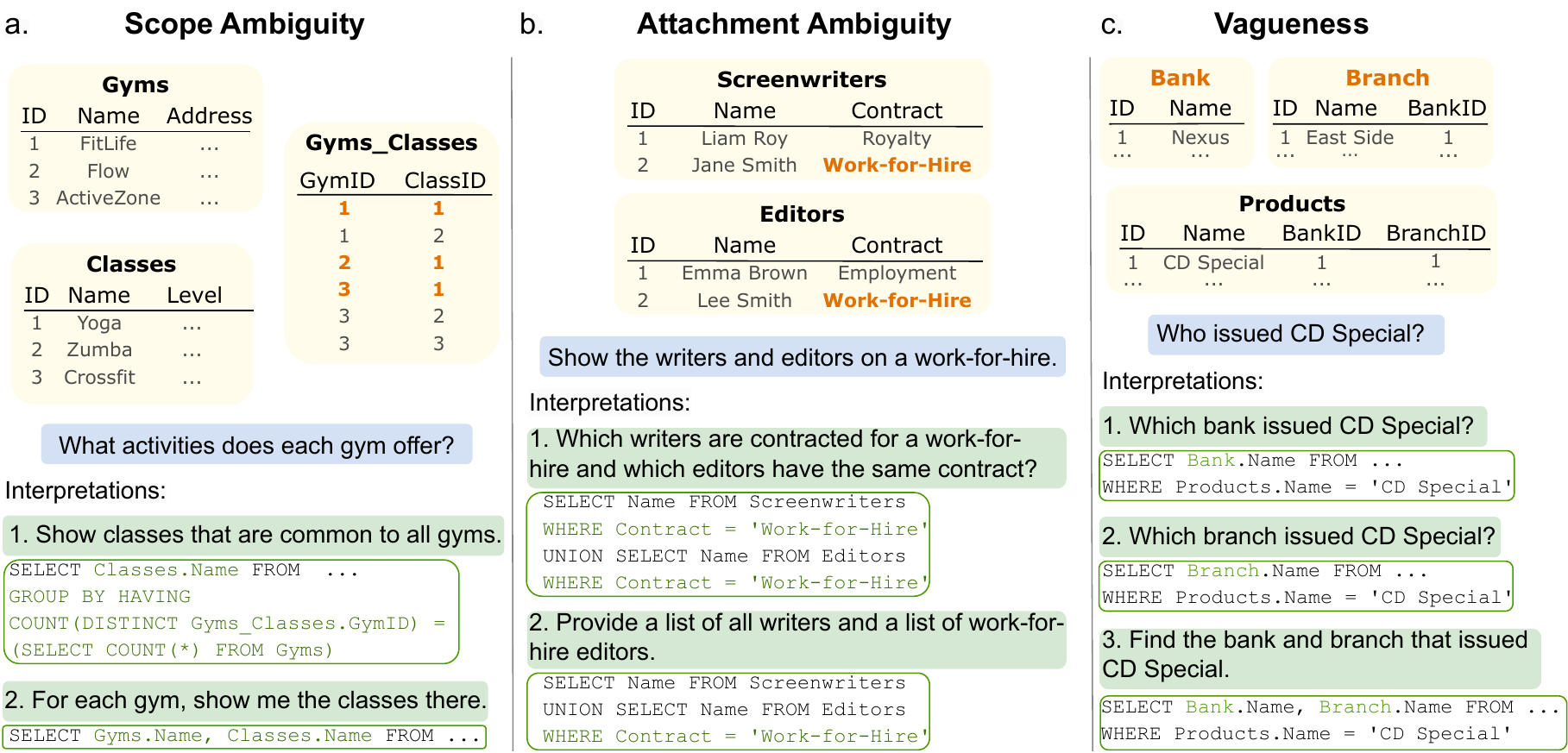}
\caption{Types of ambiguous questions (highlighted in blue), their interpretations (highlighted in green), and
  corresponding SQL queries. Database elements that could lead to ambiguity are highlighted in orange.}
\label{fig:ambig_example}
\end{figure}

Semantic parsing translates natural language utterances to logical
forms or executable programs in some machine-readable language
(e.g.,~SQL). It has emerged as an important component in many
real-world applications
\citep{Ozcan:ea:2020,Liang:2016,wang-etal-2023-mconala,dukes-2014-semeval}
as it allows users to seek information and control computer systems
naturally and flexibly in natural language. Practical semantic parsers
are expected to understand user utterances and map them to executable
forms, %
even when these are ambiguous
(see
Figure~\ref{fig:ambig_example} where a user request allows multiple
interpretations, each corresponding to a different logical
form). 
Ambiguity is a pervasive challenge in natural language
applications \citep{min-etal-2020-ambigqa, liu-etal-2023-afraid, yuan-etal-2023-ambicoref}, and semantic parsing is no exception.
\citet{wang-etal-2023-know} show that more than half of failure cases
for a text-to-SQL semantic parser are due to ambiguity which can occur
in different forms and at different levels. %

Although the problem of mapping natural language utterances to formal
representations has been studied extensively, the issue of ambiguity
has received less attention. %
\citet{eskin-2024-zero} evaluate the ability of large language
models to parse ambiguous sentences to first-order logic, focusing on five well-known linguistic ambiguities. %
In the context
of text-to-SQL parsing, other work \citep{wang-etal-2023-know,
  bhaskar-etal-2023-benchmarking} introduces vagueness into the
questions of popular benchmarks like Spider
\citep{yu-etal-2018-spider}
by modifying their databases, e.g., through synonyms.
Although targeting a real-world application, database augmentation is
limited to a single type of ambiguity\footnote{Vagueness and ambiguity
are often considered distinct properties \citep{frappier2012analysis};
however, for simplicity, we will refer to vagueness as a type of
ambiguity.}  and often operates in an artificial setting.  For
example, consider the database shown in
Figure~\ref{fig:ambig_example}b. We could add a ``Scriptwriters'' table
with the same content as the existing ``Screenwriters'' one. Our new
database would allow vague questions, but would not be very realistic
or well-designed.

In this paper we introduce \dataset{}, a novel benchmark  which we hope will both inform and inspire the development of parsers capable of 
recognizing and interpreting ambiguous queries.  \dataset{} covers 16~distinct domains, it contains
846~multi-table databases, ambiguous questions,  their unambiguous interpretations provided by humans, %
and complex SQL queries (4,242 in total). 
It  includes three types of ambiguity, i.e.,~scope
ambiguity, attachment ambiguity, and vagueness, 
showcasing a diverse range of SQL queries. 
Figure~\ref{fig:ambig_example} shows examples of  ambiguous questions in blue blocks and their unambiguous interpretations in green, both of which can be phrased as requests.
Aiming
to mimic real-world semantic parsing scenarios
with realistic and diverse databases, we create them automatically in three steps:
(a)~we specify a domain of interest (e.g.,~%
Banking); (b)~we
generate key concepts and relations such that they obey constraints
imposed by a chosen ambiguity type (e.g.,~for scope ambiguity,
the database must have a many-to-many relationship with a common element; %
see
Figure~\ref{fig:ambig_example}a where multiple gyms offer the same
class, namely yoga); and (c)~we generate SQL statements to construct
tables with the desired structure. %
We
use a large language model for generation (OpenChat; \citealt{Wang-2024-openchat})
and  view database creation as a semantic parsing
problem. Since we can automatically filter predicted SQL statements
based on execution results, minimal manual effort is required to
validate the generated databases and their content.\footnote{The code and data are publicly available at: \url{ambrosia-benchmark.github.io}}

We benchmark multiple advanced large language models of different sizes on \dataset{}, including the most recent
Llama~3~\citep{dubey2024llama3herdmodels} and GPT-4o. Our experiments reveal that models struggle to recognize ambiguity and provide all possible SQL queries for all interpretations. They often capture only one interpretation and are biased towards a specific type of ambiguity. 
The best model, Llama~3-70B, achieves only 31\%~recall on ambiguous questions compared to 66\%~on unambiguous ones.  
\dataset{} offers a diverse range of questions from various domains, each introducing distinct types of ambiguity along with their interpretations. This diversity offers invaluable insights into the challenges of real-world semantic parsing.

\section{Related Work}
\looseness=-1 
The ambiguity inherent in natural language has been studied through
the lens of various tasks, including question-answering
\citep{min-etal-2020-ambigqa}, natural language inference (NLI;
\citealt{liu-etal-2023-afraid}), and coreference resolution
\citep{yuan-etal-2023-ambicoref},
where models have been broadly found
lacking in their ability to resolve ambiguities.
The bulk of previous work has focused on question answering, with
emphasis on asking clarification questions to understand user intent
\citep{rahmani-etal-2023-survey}, open-domain question answering where
a query can plausibly have multiple valid answers
\citep{min-etal-2020-ambigqa}, disambiguating database search results
in the context of task-oriented dialogue systems
\citep{qian-etal-2022-database, kim-etal-2023-injecting}, and
leveraging relevance feedback to rerank the answers returned from a QA
engine based on knowledge graphs \citep{Lihui2023kgqa}.

Within the broader area of semantic parsing, some work
\citep{li-etal-2023-python, mu2023clarifygpt} has concentrated on
clarifying vague questions for code generation. There is also interest
in creating datasets with ambiguous utterances and corresponding
logical form representations.  \citet{rasmussen-schuler-2020-corpus}
collect a dataset of $\lambda$-calculus translations that includes
examples of scope ambiguity, \citet{arthur-etal-2015-semantic} explore
different types of ambiguities that arise in the task of mapping
search queries into SCFGs, and \citet{eskin-2024-zero} create a
benchmark for mapping ambiguous sentences into first-order logic. 
We also collect ambiguous utterances, however, our benchmark is
designed for parsing questions into SQL database queries.  Unlike
\citet{arthur-etal-2015-semantic} and \citet{eskin-2024-zero} who
create synthetic examples from templates, we ask human annotators to
write natural questions for a real-world application.

Related work in text-to-SQL parsing has primarily focused on vague questions. \citet{wang-etal-2023-know} detect questions containing ambiguous
tokens that could map to multiple columns.  Their dataset builds on
WikiSQL \citep{zhongSeq2SQL2017} and Squall
\citep{shi-etal-2020-potential} which are limited to
\emph{single-table} databases.
\citet{bhaskar-etal-2023-benchmarking}
modify Spider \citep{yu-etal-2018-spider} with ChatGPT to create
databases that exclusively support vague questions.
Despite relying on Spider, their approach often yields unrealistic
databases, e.g.,~they introduce ambiguity in table names by copying and renaming existing tables, which leads to information being
duplicated.  
Our dataset  not only supports vagueness but also includes scope and attachment ambiguities. Moreover, \dataset{} provides \emph{multi-table} databases that mirror realistic applications. 

\citet{Huang2023} explore ambiguity in the KaggleDBQA  dataset \citep{lee-etal-2021-kaggledbqa} focusing on vagueness, underspecified output formats, and unknown data structures. In contrast, our work assumes the database context is fully specified and focuses on different types of linguistic ambiguity. 
Other work \cite{Floratou2024, pourreza-rafiei-2023-evaluating} analyzes vague questions and highlights issues in existing text-to-SQL benchmarks, where ambiguous questions are often linked to only one SQL query, leading to execution accuracy failures. \dataset{} addresses this limitation by providing multiple SQL interpretations for ambiguous questions.
\citet{Veltri2023DataAP} automatically generate declarative sentences containing facts that may lead to contradictions due to vague tables. Our approach, however, centers on human-written and verified questions that users might ask in real-world scenarios, rather than fact-checking.

Overall, our dataset introduces various types of
ambiguity in questions, including scope and attachment ambiguities, which are often overlooked. Additionally, \dataset{} features human-written interpretations and diverse SQL queries within a single ambiguity
type and multi-table databases.

\section{\texorpdfstring{The $\varmathbb{AMBROSIA}$ Dataset Creation}{~}}

\subsection{Formal Definition of Ambiguity}
Before discussing how our dataset was created, we formally define ambiguity in text-to-SQL parsing, adapting the definition presented in \citet{Floratou2024}:

\begin{definition}
Two SQL queries are \textbf{non-equivalent} if they produce different execution results, notwithstanding variations in layout or format.
\end{definition}

\begin{definition} 
Let $Q = \{q_1, \dots, q_N \}$ denote the universe of non-equivalent SQL queries that can be formulated given a database $D$, with known database schema and values. Let $s$ denote a natural language question and $f: s \rightarrow P(Q)$ a function that operates in the context of database $D$ and deterministically maps $s$ to $P(Q)$, the power set of $Q$. Question~$s$ is \textbf{ambiguous} if $f(s)$ has a cardinality of at least two.
\end{definition}

This definition excludes ambiguities emanating from data management issues (e.g.,  relating to formatting, coverage, or the handling of \texttt{NULL} values), and assumes that the database schema and values are known. We also do not consider underspecification of the output format (e.g.,~whether the result should contain only specific columns or if auxiliary columns are acceptable).
Instead, our focus  is on ambiguity as a  \textbf{linguistic phenomenon}, arising from the way a question is formulated, and leading to multiple interpretations and corresponding SQL queries. This ambiguity persists because the database context does not uniquely resolve the interpretations a question invites. 

\subsection{Design Considerations \label{sec:design}}
\paragraph{Executable Logical Forms} In designing our benchmark, we   concentrated on  text-to-SQL semantic parsing for several
reasons. It represents a real-world use case where ambiguity arises naturally, e.g.,~in  questions posed by 
users and the structure of  databases.  Unlike some
other logical forms, SQL queries can be easily executed and inspected
for correctness. The task is familiar to LLMs, they have
demonstrated strong performance on standard benchmarks like Spider
\citep{yu-etal-2018-spider} and BIRD-Bench \citep{li-etal-2023-can}; it  is reasonable to expect them to be able to parse SQL queries even
in zero-shot settings since they likely have learned SQL syntax during
training. This allows us to focus on ambiguity per se, rather than the
model's ability to generate well-formed SQL. 

\paragraph{Databases that Support Ambiguity} Another important
consideration is ambiguity \emph{in the context of a database}; it is
not enough to just have ambiguous questions, they must also retain
their ambiguity in relation to the database context.  For instance,
the question ``What activities does each gym offer?'' in
Figure~\ref{fig:ambig_example}a is ambiguous precisely because there are
fitness classes in the database common to multiple gyms (see the ``Gyms\_Classes''
table). Most databases  in academic text-to-SQL benchmarks
(e.g.,~Spider) %
do not support ambiguous
questions. %
As discussed earlier, modifying
these databases, e.g.,~by adding tables or columns with synonymous
names, makes them unrealistic with duplicate
information and does not cover different types of ambiguity.
In Section~\ref{sec:database-generation} we describe a controllable,
multi-step approach that uses LLMs to generate databases supporting
question ambiguity.

\paragraph{Different Ambiguity Types}
Finally, we wish to include different types of ambiguity  (see Figure~\ref{fig:ambig_example}). 

\looseness=-1
\emph{Scope ambiguity} arises when it is unclear 
which elements a quantifier, such as ``each'', ``every'', or ``all'', refers to.
There are two possible interpretations
for the ambiguous question in Figure~\ref{fig:ambig_example}a: in the collective
interpretation, the quantifier is interpreted widely (i.e.,~``each gym''
refers to all gyms in the database) and in the
distributive interpretation the quantifier is interpreted narrowly
(i.e.,~``each gym'' is considered separately).

\looseness=-1
\emph{Attachment ambiguity} occurs when it is unclear how a modifier
or phrase is attached to the rest of the sentence. There are two
possible interpretations for the question in
Figure~\ref{fig:ambig_example}b: in the high attachment reading, the
prepositional phrase  ``on a work-for-hire'' is attached to the verb ``show'' (i.e., both
screenwriters and editors are on work-for-hire contracts), whereas in the low
attachment reading it is attached to ``editors'' (i.e.,~only editors
have work-for-hire contracts, and screenwriters are on any contract). Within this
category, we also consider attachment ambiguities for relative clauses
(e.g., ``writers and editors who have work-for-hire contracts'') and adjectives
(e.g., ``work-for-hire editors and screenwriters'') as their underlying
 database structure and SQL queries are similar to prepositional
phrases. 

\emph{Vagueness} occurs when context creates uncertainty about which
set of entities is being referred to. Similarly to ambiguous
questions, there can be several interpretations. In the example in
Figure~\ref{fig:ambig_example}c, the question has three
interpretations depending on whether the answer refers to a general
entity (e.g.,~the bank) or a more specific subtype (e.g.,~the branch),
or both.

\looseness=-1
Scope and attachment ambiguities are well-known examples of structural ambiguity \citep{resnik-1993-semantic, Kearns2000-KEAS, carnie2013syntax, Kiss2017} that arise when a sentence has more than one syntactic parse. However, the research community has only recently started exploring them in the context of LLMs \citep{liu-etal-2023-afraid, Kamath-etal-2024-scope, eskin-2024-zero}. 
We classify vagueness separately, as vague questions typically have a single syntactic parse, but, due to semantic imprecision, can refer to different database entities. 
We recognize that we do not exhaustively cover all cases of ambiguity in questions. For instance, we do not address lexical ambiguity (e.g.,~``Mississippi'' as a river vs. state) which  is less common in our context. We hope follow-on work will augment our dataset with  additional types of ambiguity.

\begin{figure}[t]
\centering
\centering
\includegraphics[width=1\textwidth]{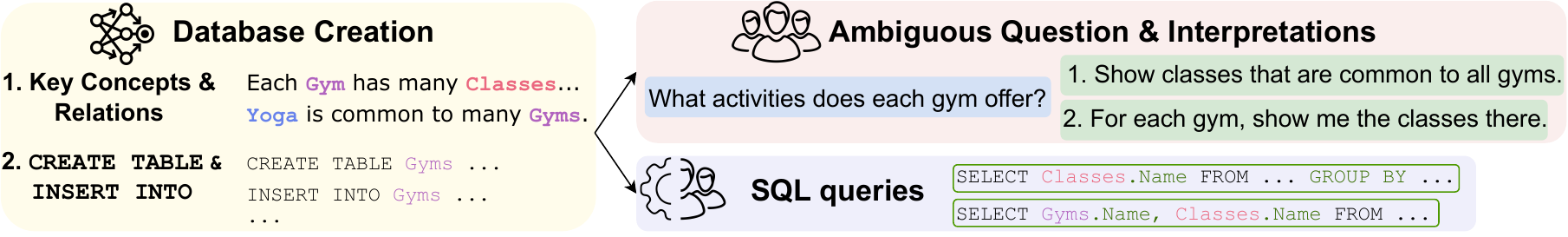}%
\caption{Annotation process for scope ambiguity  in the ``Health'' domain.}
\label{fig:annotation_process}
\end{figure}

We create \dataset{} with these considerations in mind, following 
three steps: we select a domain and generate a database that supports one of the above ambiguity types; next, human annotators write ambiguous questions along with their
unambiguous interpretations; %
finally, we automatically generate SQL queries based on templates for scope and attachment ambiguity, and ask annotators to write SQL queries for vague questions. Figure~\ref{fig:annotation_process} shows a sketch of the annotation process for scope ambiguity.

\subsection{Database Generation}
\label{sec:database-generation}
In this section we describe the database generation process. %
We break this task into the following subtasks: selecting domains,  generating key concepts and relations for each type of  ambiguity,   and generating \texttt{CREATE TABLE} and \texttt{INSERT INTO} SQL~statements to build the database.

\paragraph{Domains, Concepts, and Relations}

After analyzing existing semantic parsing datasets and publicly
available databases, we compiled a list of real-world domains for
database use. We refined this list to 16~domains (e.g.,~Entertainment,
Banking, Hospitality; see the full list in Appendix~\ref{apdx:domains}) as we found that some were either too narrow or
too broad for LLM-based database generation.

To allow different interpretations in the context of a database, the
latter must comply with constraints specific to a particular type of
ambiguity. In the scope ambiguity example in
Figure~\ref{fig:ambig_example}a, the database has information
pertaining to different gyms and classes: a gym can offer multiple
classes and a class can be offered by multiple gyms %
but one class, i.e.,~yoga, is common among them. Due to
this structure (a many-to-many relationship with a common element),
the question ``What activities does each gym offer?'' has two different
interpretations (collective vs. distributive reading).

\begin{figure}[t]
\includegraphics[width=\textwidth]{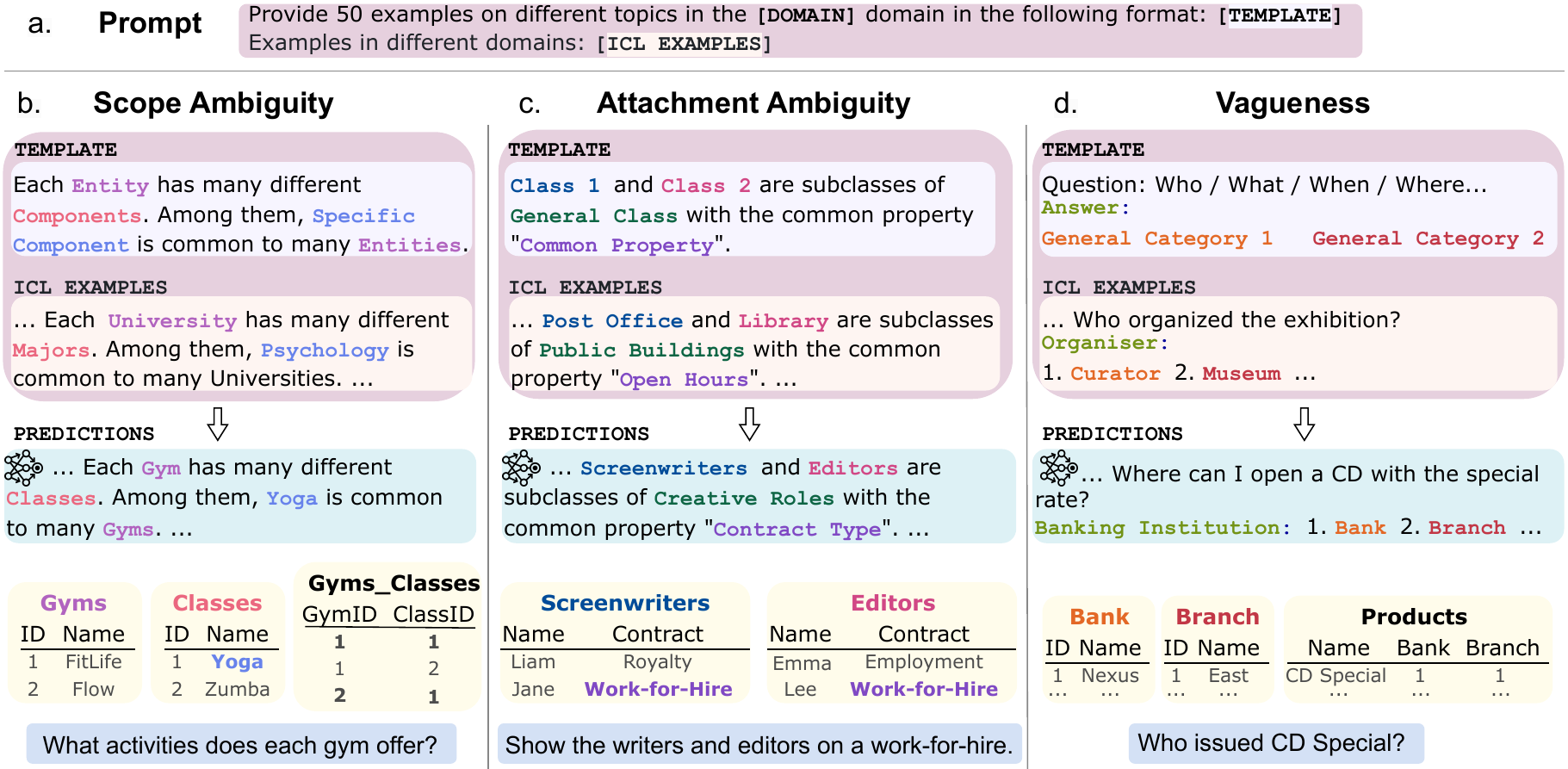}%
\caption{%
  The prompt,
  templates, in-context examples (only one out of ten is shown for brevity, see Appendix~\ref{apdx:prompts_concepts} for the full versions), and predictions of key concepts and relations for each ambiguity type. 
  Generated
  key concepts and relations later become sources of ambiguity in
  questions and databases (shown at the bottom for illustrative
  purposes).}
\label{fig:ambig_concepts}
\end{figure}

Similar to the conceptual data modeling stage used in database
modeling, we first identify  key concepts and relations that later
become sources of ambiguity in questions and databases. For each type
of ambiguity, we define a template that captures the general structure
of the data and the potential for ambiguity. Using this template and
ten in-context learning examples, we generate different structures
(key concepts and relationships) within a given
domain. Figure~\ref{fig:ambig_concepts} shows the prompt, templates,
in-context examples, and predictions which we obtain with a large
language model (see Appendix~\ref{apdx:prompts_concepts} for the full prompts).  Specifically, we employ OpenChat
\citep{Wang-2024-openchat}, one of the strongest open-sourced 7B LLMs with high performance in both code and text generation.

We manually inspect and filter LLM predictions with hallucinations or
expressions otherwise unsuitable as database elements (e.g.,~proper
nouns are unlikely to serve as table or column names). We found the
vague category to be the most difficult to generate, requiring the
most filtering.  Examples include entities unsuitable for
further use as table or column names (e.g., ``CGI'' and ``Practical
Effects'' can  be valid answers for ``What special effects were used
in a movie?'' but do not suit table or column names in a realistic
database) or mutually exclusive (e.g., ``Major Studio'' and ``Indie
Producer'' %
generally do not produce
the same movie but are both types of
production). Out of~1,883 predictions generated across all
three categories, 881 (47\%) were retained.

\paragraph{Database Generation via SQL statements}
\looseness=-1 
Based on the domain, key concepts, and relations from the
previous step, we generate and execute \texttt{CREATE TABLE} statements to
define database tables and their structure, as well as \texttt{INSERT INTO}
statements to add values to tables. For this, we use the OpenChat model in a zero-shot setting. %
We determine possible database configurations for each type of
ambiguity, since the concepts and relations generated in the previous
step can be mapped to database elements in different ways. For
instance, ``Screenwriters'' and ``Editors'' are two concepts predicted for
the attachment example in Figure~\ref{fig:ambig_concepts}c, and can be
used as table names (shown in the example) or column names within a
single table. 
In total, we
consider four configurations for databases that support attachment
ambiguity, two for databases that support vagueness, and one for scope
ambiguity, each of these corresponds to a separate instruction to the
LLM (see details in Appendix~\ref{apdx:db_generation}).

We first generate \texttt{CREATE TABLE} statements (4--6~tables per database on
average), and then proceed to generate \texttt{INSERT INTO} statements
(3--5~rows per table on average). At each step, we do not accept
predictions that are non-executable or different from the selected
configuration.  We manually validate the generated databases and
filter those we consider unnatural.  For example, in
Figure~\ref{fig:ambig_concepts}c, a database with ``Contract Type'' as a
table would not be realistic, even though possible to generate.
In
contrast, ``Projects'', an alternative common property for screenwriters and editors, could
be mapped to a column or table name.
The total number of databases
 we obtain after filtering is~846.

\subsection{Question and SQL Annotation}

Ambiguous questions and their interpretations were written by human
annotators. %
SQL queries for scope and attachment ambiguity were generated
automatically, while those for vague questions were written by annotators. Annotators were recruited via the Prolific crowdsourcing  platform based on various screening tasks; they  were all native English speakers with prior knowledge of SQL or experience in database management.
We manually reviewed all submissions and ensured all annotators followed our instructions. 
Details on crowdsourcing are provided in Appendix~\ref{apdx:annotators}.

\paragraph{Scope and Attachment Ambiguity}
\looseness=-1 
To create examples of scope and attachment ambiguity, we first
automatically generate SQL queries using pre-defined templates for
each database configuration (one for scope and four for attachment
ambiguity). SQL queries are executed to ensure they yield
non-empty, distinct results (and thus correspond to an ambiguous
intent). We  then automatically generate templates for questions (e.g.,~``What
\texttt{Name} of \texttt{Classes} do we have for every \texttt{Gyms}?'' in Figure~\ref{fig:ambig_example}a). Prolific annotators (20 in total) are
shown a database as context and 
asked to write ambiguous questions and
their interpretations in natural language based on these templates.
The annotators substantially edited and paraphrased the original templates. The mean edit distance between our templates and their
annotations is 9.2 for scope ambiguity and 12.3 for attachment ambiguity (higher values indicates greater deviation from the template, and edit distance is 0 when there is no difference).

In total, we obtained 501~questions with scope ambiguity, 362~questions with attachment ambiguity and two interpretations in natural language per question. Each ambiguous question corresponds to two golden SQL queries, and each interpretation is itself an unambiguous question and corresponds to one of these SQL queries. 

\paragraph{Vagueness}
A major difference between vague questions and those with scope and
attachment ambiguities is that the former can map to very different
SQL queries. To promote the diversity of SQL queries in our dataset,
we ask experts (10 in total) to write SQL queries from scratch for this category
rather than relying on predefined templates.  To make the task easier,
we simplify the databases by dropping and renaming tables or
columns. For example, the ``Bank'' and ``Branch'' tables shown in
Figure~\ref{fig:ambig_concepts}d are merged into a ``Banking Institution''
table (so that annotators inspect one general entity instead of two
specific ones). Annotators write SQL queries and corresponding
questions for the simpler tables (e.g.,~``Banking Institution'') which
we restore to the original databases once the annotation is
complete. A question is ambiguous as an entity or its reference can be
interpreted in different ways (e.g., a banking institution might be a
bank, a branch, or both). From the SQL query written by the
annotators, we create SQL queries corresponding to different
interpretations by replacing table or column names referring to the
general entity with the specific original names. Again, we execute the
queries to validate that they produce non-empty, distinct results.

Finally, a different batch of annotators  (the same
ones who  wrote scope and attachment questions and
interpretations) were shown vague questions and corresponding key concepts (e.g., a vague question ``Who issued CD Special?'' and key concepts ``Bank'' and ``Branch'' shown in Figure~\ref{fig:ambig_concepts}d) and asked  to write interpretations in natural language.  In total, we obtained
414 vague questions and 1,239 interpretations (two or three per
question). Each  question has  two or three
golden SQL queries, and each interpretation
is itself an unambiguous question and corresponds to one of these SQL
queries.

\begin{table}[t]
\centering
\caption{Comparison between \dataset{} and other text-to-SQL datasets with ambiguous questions, NoisySP \citep{wang-etal-2023-know} and AmbiQT \citep{bhaskar-etal-2023-benchmarking}. 
\#  ambig and \#  unambig refer to the number of ambiguous questions and their unambiguous interpretations.%
}
\label{tab:datasets}
\footnotesize

\begin{tabular}{@{}lr@{\;\;}rr@{\;\;}rr@{\;\;}rrc@{}}
& \multicolumn{2}{c}{\textbf{Scope}} & \multicolumn{2}{c}{\textbf{Attachment}} & \multicolumn{2}{c}{\textbf{Vague}} & \multirow{2}{*}{\textbf{\# DB}} & \multirow{2}{*}{\textbf{\# Tab/DB}} \\
& \textbf{\# ambig} & \textbf{\# unambig} & \textbf{\# ambig} & \textbf{\# unambig} & \textbf{\# ambig} & \textbf{\# unambig} & & \\\toprule
NoisySP & 0 & 0 & 0 & 0 & 8,673  & 0 & 8,086 & 1.0 \\
AmbiQT & 0 & 0 & 0 & 0 & 23,295 & 0 & 200 & 5.1 \\ \midrule
\dataset{} & 501 & 1,002 & 362 & 724 & 414 & 1,239 & 846 & 5.0 \\\bottomrule
\end{tabular}
\end{table}

\subsection{Dataset Analysis}
\looseness=-1
Table~\ref{tab:datasets} shows  dataset statistics for \dataset{} compared to two other text-to-SQL datasets with ambiguous questions \citep{wang-etal-2023-know,bhaskar-etal-2023-benchmarking}.  A unique aspect of \dataset{} is that it includes three different types of ambiguity and provides  interpretations in natural language for  ambiguous questions. 
The number of tables per database in \dataset{} is comparable to AmbiQT, which is based on Spider. Incidentally, our database generation approach could be used  to augment existing text-to-SQL benchmarks, e.g.,~to assess robustness or out-of-domain generalization.

\section{Experiments \label{sec:experiments}}

Below we present an experimental framework for evaluating model performance on \dataset{} and offer insights into model capabilities and failure modes. Implementation details are provided in Appendix~\ref{apdx:implem_details}, and additional experimental results can be found in Appendix~\ref{apdx:additional_results}.

\paragraph{Models} We benchmark various large language models (LLMs) on \dataset{}  in light of their growing use and good performance on text-to-SQL tasks \citep{yu-etal-2018-spider,li-etal-2023-can}. Our experiments include LLMs of different sizes: \textbf{OpenChat-7B}~\citep{Wang-2024-openchat}, the model  we used for database generation; the \textbf{instruction-tuned Llama3-8B} and \textbf{Llama3-70B} models from the Llama family \citep{dubey2024llama3herdmodels};  \textbf{instruction-tuned CodeLlama-70B}~\citep{roziere2023code} which is trained specifically for code generation tasks; and \textbf{GPT-3.5 Turbo} and \textbf{GPT-4o} models from OpenAI.

\paragraph{Prompting} 
\dataset{} offers various options for exploring ambiguity in semantic parsing. 
In general, we expect a performant model  to be able to recognize ambiguity in the context of a database and output as many SQL interpretations as applicable. Thus,
our experiments  follow two  scenarios: (1)~the model is given  instructions that acknowledge the  potential of ambiguity in the questions and specify that the output should include SQL queries for each possible interpretation; and (2)~we provide standard text-to-SQL instructions %
but consider  top-5 predictions from a beam of size~5 as does previous work \citep{bhaskar-etal-2023-benchmarking, eskin-2024-zero}.
We refer to the first method as \textbf{Prompt} and the second as \textbf{Beam}.
In both scenarios, models have access to  database context as we  display the \texttt{CREATE TABLE} and \texttt{INSERT INTO} statements which fully describe the database schema and content. Databases in \dataset{} do not have many rows, and as such fit within the context limits of the LLMs we use. We acknowledge that in real-world applications database content can be very large, requiring specific methods to extract  related database entities. However, we leave this to future work. 
For Prompt, we conduct experiments with temperature equal to 0.5 and 5 random seeds and average the results for all models except for the OpenAI ones (due to cost constraints). For Beam, experimental results are deterministic since  temperature is fixed to~0.  
We reserve 10\% of the dataset for few-shot learning.

\paragraph{Evaluation Metrics}
A common approach to evaluating text-to-SQL semantic parsing is to compare whether the predicted SQL retrieves the same answer from the database as the gold logical form, typically by measuring execution accuracy. This method accommodates different formulations of the same SQL query and we employ it as well, but in our case, the output for ambiguous questions can be mapped to \emph{several} correct SQL queries. We report recall and precision, but 
our primary focus is \emph{recall} on \emph{ambiguous} questions, as it captures the extent to which a model predicts different SQL queries that correspond to all possible interpretations. 
Following recent work \citep{bhaskar-etal-2023-benchmarking, eskin-2024-zero}, we also
measure whether \emph{all} SQL queries are generated for ambiguous questions, i.e.,~whether recall is 100\%. We call this metric AllFound. 

\subsection{Zero-Shot Results \label{sec:zero-shot}}

\begin{table}[t]
\centering
\caption{Recall, precision and AllFound metrics  for different zero-shot LLMs on \dataset{}. Llama3-70B (Prompt) captures ambiguity best, with the highest recall on ambiguous questions. All models are unable to parse multiple interpretations when these exist (see AllFound metric).}
\label{tab:zero_shot}
\footnotesize
\begin{tabular}{lcccccc}
\multirow{2}{*}{\textbf{Model}} & \multirow{2}{*}{\textbf{Method}} & \multicolumn{2}{c}{\textbf{\% Recall}} & \multicolumn{2}{c}{\textbf{\% Precision}} & \multicolumn{1}{c}{\textbf{\% AllFound}} \\
& & \textbf{ambig} & \textbf{unambig} & \textbf{ambig}& \textbf{unambig}& \textbf{ambig}\\\toprule
\multirow{2}{*}{OpenChat-7B} & Prompt & 15.5  & 36.8  & 24.7  & 28.2  & 0.2  \\
 & Beam & 14.7  & 37.9  & %
 --- & --- 
 & 1.1 \\
\multirow{2}{*}{Llama3-8B} & Prompt & 18.0  & 45.4  & 30.2  & 37.9  & 0.1 %
\\
 & Beam & 19.9 & 48.6
  & --- & --- 
 & 1.7 \\
\multirow{2}{*}{CodeLlama-70B} & Prompt
& 17.9 & 44.1 & 34.3 & 40.9 & 0.1 \\
 &  Beam & 25.4 & 56.2
   & --- & --- 
& 0.1  \\
\multirow{2}{*}{Llama3-70B} & Prompt & \textbf{30.7} & 64.5  & 42.7 & 49.4 & \textbf{1.9} \\
 & Beam & 28.0 & \textbf{65.5}
   & --- & --- 
 & 1.4 \\ \hline
GPT-3.5 Turbo  & Prompt & 26.7 & 61.6  & 40.2  & 52.1  & 0.5  \\
GPT-4o & Prompt & 27.1  & 63.4  & \textbf{51.1} & \textbf{59.6} & 0.4 \\ \bottomrule
\end{tabular}
\end{table}

Table~\ref{tab:zero_shot} summarizes the performance of zero-shot models on \dataset{}. We report results %
using micro-averaging. The standard deviation of the Prompt method is within~0.3\%--1.2\% for precision and recall for all models, except for CodeLlama, which varies from~4\% to~5\%. The standard deviation for AllFound is below~0.3\%.
For fairness, we do not report precision  for the Beam method as
 it consistently outputs the top-5 predictions, although there are only 1--3~gold SQL queries.

As can be seen, all models demonstrate substantially higher recall on unambiguous questions compared to ambiguous ones,  with differences ranging from 21\% for OpenChat-7B to 36\%~for GPT-4o and 38\%~for Llama3-70B.  As indicated by the AllFound metric, models generally fail to capture the ambiguity in the question as they rarely predict SQL queries for different interpretations.
They often predict a correct SQL query for  one interpretation only, which is why precision on ambiguous questions is higher than recall. Conversely,  models sometimes predict more than one SQL query for unambiguous questions, which explains  why they have  lower precision than recall. Precision is 1.0 for both ambiguous and unambiguous questions when a model produces a \emph{single} correct interpretation. However, when a model produces multiple predictions and only one is correct precision can drop significantly (e.g., to~0.1). This effect explains why ambiguous and unambiguous questions obtain somewhat similar precision.

Note that recall for unambiguous questions is the same as execution accuracy in standard text-to-SQL with one gold SQL query. Table~\ref{tab:zero_shot} shows that the best recall is only 65.5\% (achieved by Llama3-70B), which suggests \dataset{} has challenging examples even in a standard text-to-SQL setting.

Overall, Llama3-70B (Prompt) captures ambiguity best, with the highest recall on ambiguous questions at nearly 31\% and the highest AllFound value of 1.9\%  which is admittedly still very low. Llama3-70B performs better with the Prompt method, while CodeLlama-70B appears to perform better when considering top-k predictions (Beam),  however, it never predicts both interpretations in this case.   CodeLlama-70B is very unstable with the Prompt method, %
and shows performance comparable to Llama3-8B.
This instability indicates that parsing ambiguous questions is significantly different from other code generation tasks encountered during training, requiring different skills (e.g.,~explicit instructions or documentation).
GPT-4o performs best in terms of precision, however,  even for this model, precision is lower than recall on unambiguous questions which means that it predicts SQL queries when it should not. \textbf{In general, all models fail to provide multiple SQL queries when several interpretations are possible due to ambiguity  but can mistakenly offer more than one SQL query for unambiguous questions.} %

\subsection{Analysis}
\begin{table}[t]
\caption{Breakdown of model performance (zero-shot Llama3-70B)  by ambiguity type shows that attachment ambiguity is most challenging.}
\label{fig:llama_analysis}
\footnotesize
\begin{tabular}{lll}
\begin{minipage}[t]{2in}
\hspace*{-.2cm}\begin{tabular}{@{}l@{}c@{~}c@{~~}c@{\hspace*{-.2cm}}c@{\hspace*{-.5cm}}c@{\hspace*{-.3cm}}}
             & \multicolumn{2}{c}{\bf \% Recall} &
  \multicolumn{2}{c}{\bf \% Precision} & \multicolumn{1}{c@{}}{\bf \% AllFound} \\
{Prompt}        &  \textbf{ambig} & \textbf{unambig} & \textbf{ambig} &
  \textbf{unambig} & \textbf{ambig} \\  \toprule
  Scope       &  41.5  & 90.4  & 52.7 & 66.4 & 2.9 \\
  Attachment &   12.7  & 24.0  & 13.7 & 13.4 & 0.3\\
  Vague      &   35.6   & 69.4 & 56.7 & 56.4 & 4.6 \\ \bottomrule
\end{tabular}
\end{minipage}  & \hspace{2.5cm} & 
\begin{minipage}[t]{2in}
\begin{tabular}{@{}l@{~}c@{~}c@{}c@{\hspace*{-.2cm}}}
             & \multicolumn{2}{c}{\bf \% Recall} &  \multicolumn{1}{@{\hspace*{-.1cm}}c@{}}{\bf \% AllFound} \\
{Beam}        &  \textbf{ambig} & \textbf{unambig} & \textbf{ambig} \\  \toprule
  Scope  &  41.6  & 91.8  & 1.1 \\
  Attachment &   10.3  & 22.2   & 0.0\\
  Vague     &   25.8   & 69.1  & 0.3 \\ \bottomrule
\end{tabular}
\end{minipage}
\end{tabular}
\end{table}

We perform a more detailed analysis on the zero-shot instruction-tuned Llama 3-70B model, which has demonstrated the highest performance on our dataset.
Table~\ref{fig:llama_analysis} presents recall, precision, and AllFound metrics for different categories of ambiguous and unambiguous questions. We observe that attachment ambiguity is most challenging. This category involves complex SQL queries (often requiring the \texttt{UNION} operator) and has the largest diversity in database configurations and corresponding gold SQL queries, which explains poor performance even on  unambiguous questions. In contrast, we obtain best results for scope ambiguity, achieving nearly 92\% recall (i.e.,~execution accuracy) on unambiguous questions,  and 42\% on ambiguous ones.
We consider only one database configuration for scope ambiguity, which might be more familiar to LLMs due to the widespread use of many-to-many relationships.   Interestingly, the model captures vagueness better with the Prompt method as evidenced by superior recall and AllFound results.

Table~\ref{tab:diff_interpretations} shows the distribution of model predictions by interpretation type. We consider ambiguous examples that have at least one correctly predicted SQL query. For vague questions, we focus on those with three interpretations and define two types of predictions: those with only one component (interpretations 1 and 2 in Figure~\ref{fig:ambig_example}c; Component) and those with all components (interpretation~3 in Figure~\ref{fig:ambig_example}c; Full). 
There is a clear bias towards one interpretation type.  The distributive interpretation is preferred for scope ambiguity (in 80\% of cases) corroborating the  findings of  \citet{kamath2024scope}.  High attachment is  chosen in more than 97\%~of cases for attachment ambiguity, and interpretations involving individual components are also overwhelmingly preferred in the case of vagueness. Overall, the Prompt method leads to more biased predictions compared to Beam.

\begin{table}[t]
\centering
\caption{Distribution of model predictions (zero-shot Llama3-70B) by interpretation type. Across ambiguities, there is a clear bias towards one interpretation type.}
\label{tab:diff_interpretations}
\footnotesize
\begin{tabular}{lcccccc}%
& \multicolumn{2}{c}{\textbf{Scope}} & \multicolumn{2}{c}{\textbf{Attachment}} & \multicolumn{2}{c}{\textbf{Vague}} \\
{Method} & \textbf{Collective} & \textbf{Distributive} & \textbf{High}& \textbf{Low}& \textbf{Component} & \textbf{Full} \\\toprule
Prompt & 16.5  & 83.5  & 98.2 & 1.8  & 83.5 & 16.5 \\
Beam & 18.6  & 81.4  & 97.3  & 2.7  & 74.3  & 25.7 \\\bottomrule
\end{tabular}
\end{table}

\subsection{Few-Shot Results \label{sec:few-shot}}
\begin{figure}[t]
\centering
\includegraphics[width=\textwidth]{img/few_shot_results_camera_ready.png}

\caption{Recall, precision, and AllFound metrics for zero-shot and few-shot Llama3-70B. 
In-context examples are selected randomly. We obtain best results with  1-3 examples.}
\label{fig:few_shot}

\end{figure}

Figure~\ref{fig:few_shot} shows the performance of Llama3-70B (Prompt) in a few-shot learning setting. %
We select in-context examples randomly,  each including ambiguous questions, their unambiguous interpretations, and corresponding SQL queries. 
We observe largest improvements in recall and precision over the zero-shot method with one to three examples. Recall on ambiguous questions improves by~4\% but remains substantially worse compared to unambiguous questions. 
Increasing the number of examples helps, but  improvements are not statistically significant given the 2--7\% standard deviation.

\looseness=-1
Table~\ref{tab:few_shot_exploration}
 presents results with GPT-4o in  one-shot setting (with 3 seeds).
 As can be seen, LLaMa3-70B and GPT-4o perform similarly: one-shot improves recall and AllFound in  ambiguous questions for both models.
 The only difference is a slight (not statistically significant) decrease for LLaMa3-70B in precision, which could be due to the model generating more SQL queries, leading to more incorrect predictions and thus lower precision.

Further analysis indicates that providing more examples for one type of ambiguity improves performance for that type but may negatively impact others. %
 We thus benchmark Llama3-70B with a prompt that includes \emph{all} three ambiguities and their definitions  from Section~\ref{sec:design}, unlike random sampling. The two  prompt formats yield similar results, however, we observe  a smaller standard deviation (2–4\%) when prompting with all three ambiguities.
Since it is unrealistic to have examples for all possible ambiguities, we consider these results an upper bound.

\begin{table}[t]
\centering
\caption{
Recall, precision, and AllFound metrics  for Llama3-70B and GPT-4o in zero- and one-shot settings. We also include results for Llama3-70B  simultaneously prompted with all three ambiguities  and their definitions. Differences between models and settings are negligible.}

\label{tab:few_shot_exploration}
\footnotesize
\begin{tabular}{lcccccc}
\multirow{2}{*}{\textbf{Model}} & \multirow{2}{*}{\textbf{ICL Examples}} & \multicolumn{2}{c}{\textbf{\% Recall}} & \multicolumn{2}{c}{\textbf{\% Precision}} & \multicolumn{1}{c}{\textbf{\% AllFound}} \\
& & \textbf{ambig} & \textbf{unambig} & \textbf{ambig}& \textbf{unambig}& \textbf{ambig}\\\toprule
\multirow{3}{*}{Llama3-70B} & 0-shot & 30.7 & 64.5  & 42.7 & 49.4 & 1.9   \\
 & 1-shot & 32.4  & \textbf{67.3} & 
\textbf{51.7}  & 57.1
 & 3.7 \\
  & 3 ambiguities & \textbf{35.0}  & 66.6  & 
48.3 & 50.7
 & 3.0 \\\midrule
\multirow{2}{*}{GPT-4o} &  0-shot  & 27.1  & 63.4  & 51.1 & \textbf{59.6} & 0.4  \\
 & 1-shot & 31.3  & 63.8  & 49.8
& 53.1  & \textbf{4.5} \\
\end{tabular}
\end{table}

\section{Limitations \label{sec:limitations}} %
\looseness=-1
Despite our best efforts to create a high-quality dataset, we cannot guarantee that \dataset{} is error-free. Recall that we rely on annotators to provide ambiguous questions and their interpretations, both of which may have flaws. Hence, some interpretations may be unclear, failing to disambiguate the question, or unnatural and overly explicit with direct mentions to  database entities. 
Our databases generally have simple and clear names, whereas in reality, they might be incomplete, have %
abbreviations, and so on. Since our experiments show that LLMs struggle to detect ambiguity and provide interpretations, we believe the current databases are well-suited for our task. However, future work might include augmentations to render them more realistic. When conducting experiments, we display the full database content, which is neither scalable nor safe for real-world applications. %
Consequently, our results can be seen as an upper bound on semantic parsing performance with ambiguous questions. Although our work broadens the scope of linguistic ambiguity in the text-to-SQL task, we acknowledge our dataset does not exhaustively cover all cases of ambiguity. We hope follow-on work will explore these further.

\section{Conclusion}

In this paper, we present \dataset{}, a novel dataset for parsing  ambiguous questions into SQL database queries across multiple domains.  
We populate \dataset{} with multi-table realistic databases that support ambiguity, having  developed an automatic pipeline for  controlled database generation using key concepts and relations. \dataset{} covers three types of ambiguity and contains ambiguous questions along with their interpretations in natural language.
Our experiments demonstrate that even the most advanced LLMs struggle to capture ambiguity and provide accurate SQL queries for different interpretations of ambiguous questions, leaving ample room for improvement. We further hope \dataset{} will spur future research on generalization (e.g., across   domains and ambiguity types). Databases with fixed structures can be also modified to explore other interesting scenarios, including cases where the database context helps clarify originally  ambiguous questions.

\section{Acknowledgments}
We thank the anonymous reviewers for their constructive feedback and Tom Hosking for his insightful comments.
We gratefully acknowledge the support of the UK
Engineering and Physical Sciences Research Council (grant EP/W002876/1).

\bibliographystyle{acl_natbib}
\bibliography{anthology,custom}

\section*{Checklist}

\begin{enumerate}

\item For all authors...
\begin{enumerate}
  \item Do the main claims made in the abstract and introduction accurately reflect the paper's contributions and scope?
    \answerYes{}
  \item Did you describe the limitations of your work?
    \answerYes{See Section~\ref{sec:limitations}.} 
  \item Did you discuss any potential negative societal impacts of your work?
    \answerYes{See Section~\ref{sec:limitations}.}
  \item Have you read the ethics review guidelines and ensured that your paper conforms to them?
    \answerYes{}
\end{enumerate}

\item If you are including theoretical results...
\begin{enumerate}
  \item Did you state the full set of assumptions of all theoretical results?
\answerNA{}
	\item Did you include complete proofs of all theoretical results?
\answerNA{}
\end{enumerate}

\item If you ran experiments (e.g. for benchmarks)...
\begin{enumerate}
  \item Did you include the code, data, and instructions needed to reproduce the main experimental results (either in the supplemental material or as a URL)?
    \answerYes{}
  \item Did you specify all the training details (e.g., data splits, hyperparameters, how they were chosen)?
 \answerYes{See Section~\ref{sec:experiments}.} 
	\item Did you report error bars (e.g., with respect to the random seed after running experiments multiple times)?
 \answerYes{See Section~\ref{sec:zero-shot} and~\ref{sec:few-shot}.} 
	\item Did you include the total amount of compute and the type of resources used (e.g., type of GPUs, internal cluster, or cloud provider)?  
 \answerYes{See Appendix~\ref{apdx:implem_details}.} 
\end{enumerate}

\item If you are using existing assets (e.g., code, data, models) or curating/releasing new assets...
\begin{enumerate}
  \item If your work uses existing assets, did you cite the creators?
    \answerYes{}
  \item Did you mention the license of the assets?
    \answerYes{See Appendix~\ref{apdx:hosting}.}
  \item Did you include any new assets either in the supplemental material or as a URL?
    \answerYes{}
  \item Did you discuss whether and how consent was obtained from people whose data you're using/curating?
    \answerYes{See Appendix~\ref{apdx:annotators}.}
  \item Did you discuss whether the data you are using/curating contains personally identifiable information or offensive content?
   \answerYes{See Appendix~\ref{apdx:annotators}.}
\end{enumerate}

\item If you used crowdsourcing or conducted research with human subjects...
\begin{enumerate}
  \item Did you include the full text of instructions given to participants and screenshots, if applicable?
    \answerYes{See Appendix~\ref{apdx:annotators} and Appendix~\ref{apdx:datasheet}.}
  \item Did you describe any potential participant risks, with links to Institutional Review Board (IRB) approvals, if applicable?
    \answerNA{}
  \item Did you include the estimated hourly wage paid to participants and the total amount spent on participant compensation?
    \answerYes{See Appendix~\ref{apdx:annotators}.} 
\end{enumerate}

\end{enumerate}

\newpage
\appendix

\section{Hosting, Licensing and Maintenance\label{apdx:hosting}}
The data, instructions, and code used for the creation of \dataset{} and our benchmarking experiments are publicly available under the CC BY 4.0 license: \url{ambrosia-benchmark.github.io}.
We plan to update the dataset by correcting any annotation errors as they are identified. 
A  datasheet \citep{datasheets} for \dataset{} provides detailed documentation in Appendix~\ref{apdx:datasheet}.

\section{Domains\label{apdx:domains}}
The dataset includes the following domains across all categories of ambiguity: Airport, Banking, Education, Entertainment, Filmmaking, Hospitality, Job Postings, and Journalism. Additionally, the Scope and Vague categories include the domains of Agriculture, Traffic, Streaming Services, Healthcare, and the Housing Market. The Vague category alone also encompasses the domains of Construction, Demographics, and Students.

\section{Prompts for Key Concepts and Relations Generation\label{apdx:prompts_concepts}}
Below, we provide the prompts used for generating key concepts and relations. The domain is indicated in grey to be filled in with a specific name. We sample with a temperature of 0.6 and top\_p of 0.95 until at least 35 concepts and relations are generated, or the number of attempts exceeds 5.

    \begin{tcolorbox}[colback=white, colframe=purple, left=2pt,  coltitle=black, title=\textbf{Scope}: Key Concepts and Relations Generation, breakable, halign=flush left]
    \footnotesize
Provide 50 examples on different topics in the \textcolor{gray}{\texttt{DOMAIN}} domain in the following format:\\ \vspace{3mm}
Each \texttt{[Entity]} has many different \texttt{[Components]}. Among them, \texttt{[Specific Component]} is common to many \texttt{[Entities]}.\\\vspace{3mm}
Examples in different domains:\\
\begin{enumerate}[leftmargin=15pt]
    \item Each \texttt{University} has many different \texttt{Majors}. Among them, \texttt{Psychology} is common to many \texttt{Universities}.
    \item Each \texttt{Report} has many different \texttt{Sections}. Among them, \texttt{Introduction} is common to many \texttt{Reports}.
    \item Each \texttt{Hospital} has many different \texttt{Amenities}. Among them, \texttt{Waiting Room} is common to many \texttt{Hospitals}.
\item Each \texttt{Musical} has many different \texttt{Roles}. Among them, \texttt{Narrator} is common to many \texttt{Musicals}.
\item Each \texttt{Phone} has many different \texttt{Features}. Among them, \texttt{Touchscreen} is common to many \texttt{Phones}.
\item Each \texttt{Museum} has many different \texttt{Events}. Among them, \texttt{Family Day} is common to many \texttt{Museums}.
\item Each \texttt{Company} has many different \texttt{Departments}. Among them, \texttt{Human Resources} is common to many \texttt{Companies}.
\item Each \texttt{Website} has many different \texttt{Pages}. Among them, \texttt{Homepage} is common to many \texttt{Websites}.
\item Each \texttt{Restaurant} has many different \texttt{Dishes}. Among them, \texttt{Pizza} is common to many \texttt{Restaurants}.
\item Each \texttt{Route} has many different \texttt{Stops}. Among them, \texttt{Transit Hub} is common to many \texttt{Routes}.\\
\end{enumerate}
Examples in the \textcolor{gray}{\texttt{DOMAIN}} domain:
    \end{tcolorbox}

        \begin{tcolorbox}[colback=white, colframe=purple, coltitle=black, parskip=5mm, title=\textbf{Attachment}: Key Concepts and Relations Generation, breakable, halign=flush left]
    \footnotesize
Provide 50 examples on different topics in the \textcolor{gray}{\texttt{DOMAIN}} domain in the following format:\\\vspace{3mm}
\texttt{[Class 1]} and \texttt{[Class 2]} are subclasses of \texttt{[General Class]}. All \texttt{[Entities of Class 1]} and \texttt{[Entities of Class 2]} have property "\texttt{[Common Property]}". There might be a \texttt{[Entity of Class 1]} and a \texttt{[Entity of Class 2]} that both have "\texttt{[Common Property]}" equal to "\texttt{[Common Value]}".\\\vspace{3mm}
Examples in different domains:\\
\begin{enumerate}[leftmargin=15pt]
\item \texttt{Post Office} and \texttt{Library} are subclasses of \texttt{Public Buildings}. All \texttt{Post Offices} and \texttt{Libraries} have the property "\texttt{Open Hours}". There might be a \texttt{Post Office} and a \texttt{Library} that both have "\texttt{Open Hours}" equal to "\texttt{8 a.m.}".
\item  \texttt{Teacher} and \texttt{Lawyer} are subclasses of \texttt{Professional Occupations}. All \texttt{Teachers} and \texttt{Lawyers} have the property "\texttt{Education Level}". There might be a \texttt{Teacher} and a \texttt{Lawyer} that both have "\texttt{Education Level}" equal to "\texttt{Master's Degree}".
\item \texttt{Ballet} and \texttt{Musical} are subclasses of \texttt{Performing Arts}. All \texttt{Ballets} and \texttt{Musicals} have the property "\texttt{Performance Venue}". There might be a \texttt{Ballet} and a \texttt{Musical} that both have "\texttt{Performance Venue}" equal to "\texttt{Broadway Theater}".
\item \texttt{Apartment} and \texttt{Townhouse} are subclasses of \texttt{Residences}. All \texttt{Apartments} and \texttt{Townhouses} have the property "\texttt{Living Space Features}". There might be an \texttt{Apartment} and a \texttt{Townhouse} that both have "\texttt{Living Space Features}" equal to "\texttt{Balcony}".
\item \texttt{Bus} and \texttt{Train} are subclasses of \texttt{Public Transport}. All \texttt{Buses} and \texttt{Trains} have the property "\texttt{Scheduled Routes}". There might be a \texttt{Bus} and a \texttt{Train} that both have "\texttt{Scheduled Routes}" equal to "\texttt{Downtown Loop}".
\item \texttt{Hospital} and \texttt{Ambulatory Surgery Center} are subclasses of \texttt{Healthcare Facilities}. All \texttt{Hospitals} and \texttt{Ambulatory Surgery Centers} have the property "\texttt{Number of Patients}". There might be a \texttt{Hospital} and an \texttt{Ambulatory Surgery Center} that both have "\texttt{Number of Patients}" equal to "\texttt{Over 100 daily}".
\item  \texttt{Thesis} and \texttt{Report} are subclasses of \texttt{Academic Writing}. All \texttt{Theses} and \texttt{Reports} have the property "\texttt{Research Focus}". There might be a \texttt{Thesis} and a \texttt{Report} that both have "\texttt{Research Focus}" equal to "\texttt{Renewable Energy}".
\item \texttt{Smartphone} and \texttt{Tablet} are subclasses of \texttt{Electronic Devices}. All \texttt{Smartphones} and \texttt{Tablets} have the property "\texttt{Charger Type}". There might be a \texttt{Smartphone} and a \texttt{Tablet} that both have "\texttt{Charger Type}" equal to "\texttt{USB-C}".
\item \texttt{Documentary} and \texttt{Animation} are subclasses of \texttt{Film Genres}. All \texttt{Documentaries} and \texttt{Animations} have the property "\texttt{Target Audience}". There might be a \texttt{Documentary} and an \texttt{Animation} that both have "\texttt{Target Audience}" equal to "\texttt{Young Adults}".
\item  \texttt{Passport} and \texttt{Driving License} are subclasses of \texttt{Identification Documents}. All \texttt{Passports} and \texttt{Driving Licenses} have the property "\texttt{Date of Issue}". There might be a \texttt{Passport} and a \texttt{Driving License} that both have "\texttt{Date of Issue}" equal to "\texttt{June 1, 2018}".\\
\end{enumerate}
Examples in the \textcolor{gray}{\texttt{DOMAIN}} domain:
\end{tcolorbox}

\begin{tcolorbox}[colback=white, colframe=purple, left=2pt,  coltitle=black, title=\textbf{Vague}: Key Concepts and Relations Generation, breakable, halign=flush left]
    \footnotesize
Provide 50 examples on different topics in the \textcolor{gray}{\texttt{DOMAIN}} domain in the following format:\\\vspace{3mm}
Question: [Who / What / How / When / Where ...]? \qquad
Subject of Inquiry: \texttt{[Subject]}\\
Focus: \texttt{[Focus]} \quad
Possible answer types: 
1. \texttt{[General Category 1]} \;
2. \texttt{[General Category 2]}
\\\vspace{3mm}
Examples in different domains:\\
\begin{enumerate}[leftmargin=15pt]
\item Question: Who organized the exhibition? \qquad
Subject of Inquiry: \texttt{Exhibitions}\\
Focus: \texttt{Organizer} \qquad
Possible answer types:
1. \texttt{Curator} \;
2. \texttt{Museum}

\item Question: When was this artefact created? \qquad
Subject of Inquiry: \texttt{Artefacts}\\
Focus: \texttt{Timeframe} \qquad
Possible answer types:
1. \texttt{Historical Period} \;
2. \texttt{Estimated Age}

\item Question: What is the capacity of this venue? \qquad
Subject of Inquiry: \texttt{Venues}\\
Focus: \texttt{Capacity} \qquad
Possible answer types:
1. \texttt{Seating Capacity} \;
2. \texttt{Total Capacity}

\item Question: How was this research funded? \qquad
Subject of Inquiry: \texttt{Research}\\
Focus: \texttt{Funding} \qquad
Possible answer types:
1. \texttt{Grant} \;
2. \texttt{Funding Organization}

\item Question: Where was this play published? \qquad
Subject of Inquiry: \texttt{Books}\\
Focus: \texttt{Place of Origin} \qquad
Possible answer types:
1. \texttt{Magazine} \;
2. \texttt{Country}

\item Question: What powers this vehicle? \qquad
Subject of Inquiry: \texttt{Vehicles}\\
Focus: \texttt{Propulsion Method} \qquad
Possible answer types:
1. \texttt{Engine Type} \;
2. \texttt{Energy Source}

\item Question: When was this phone introduced? \qquad
Subject of Inquiry: \texttt{Electronics}\\
Focus: \texttt{Timeframe} \qquad
Possible answer types:
1. \texttt{Presentation Day} \;
2. \texttt{Release Date}

\item Question: Who won this contest? \qquad
Subject of Inquiry: \texttt{Contests}\\
Focus: \texttt{Winner} \qquad
Possible answer types:
1. \texttt{Singer} \;
2. \texttt{Nation}

\item Question: What are the main features of this region? \qquad
Subject of Inquiry: \texttt{Regions}\\
Focus: \texttt{Features} \qquad
Possible answer types:
1. \texttt{Landscape} \;
2. \texttt{Climate}

\item Question: Where is this painting displayed? \qquad
Subject of Inquiry: \texttt{Paintings}\\
Focus: \texttt{Location} \qquad
Possible answer types:
1. \texttt{Gallery} \;
2. \texttt{City}\\
\end{enumerate}
Examples in the \textcolor{gray}{\texttt{DOMAIN}} domain:
    \end{tcolorbox}

\section{Database Generation\label{apdx:db_generation}}
\begin{figure}[h]
\centering
\includegraphics[width=1.\textwidth]{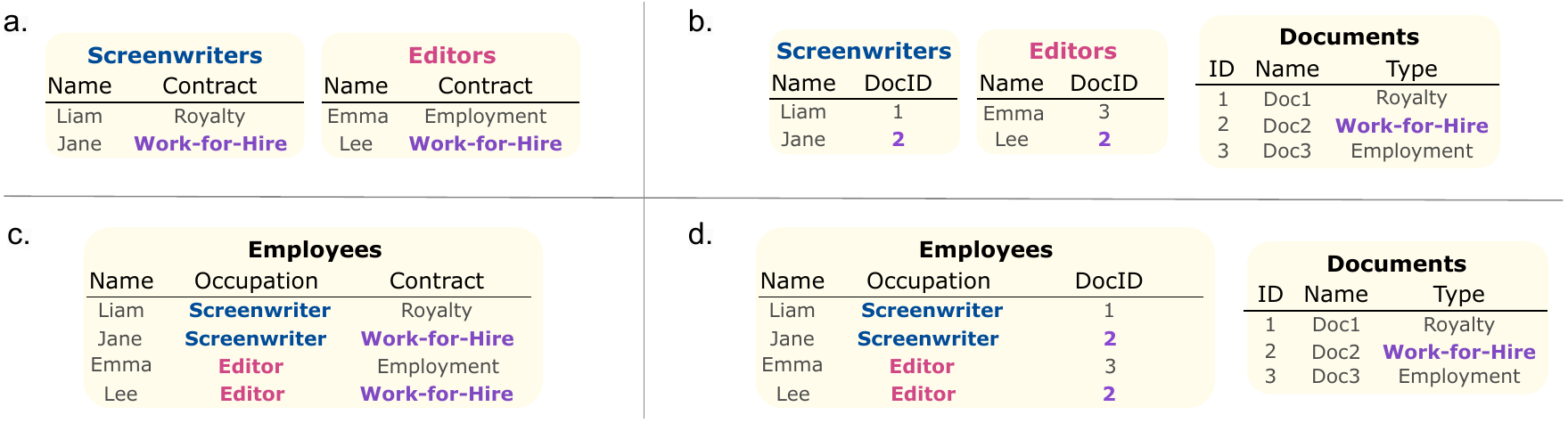}%
\caption{Database configurations that support attachment ambiguity.}
\label{fig:dbs_attach}
\end{figure}
\begin{figure}[h]
\centering
\includegraphics[width=0.75\textwidth]{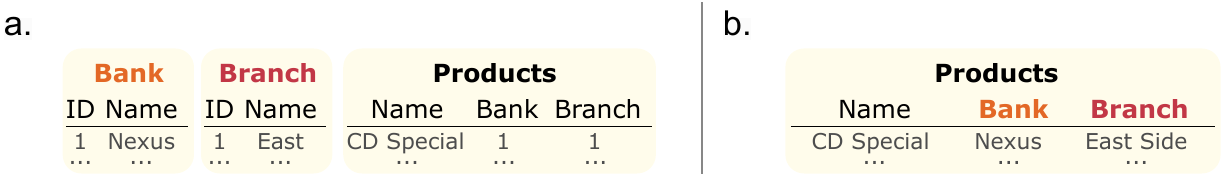}%
\caption{Database configurations that support vagueness.}
\label{fig:dbs_vague}
\end{figure}

Figure~\ref{fig:dbs_attach} shows four possible configurations for attachment ambiguity, each of them corresponding to concepts in Figure~\ref{fig:ambig_concepts}c and supporting the ambiguous request ``Show the writers and editors on  work for hire''. Figure~\ref{fig:dbs_vague} shows two possible configurations for vague questions,  each of them corresponding to concepts in Figure~\ref{fig:ambig_concepts}d and supporting the ambiguous request ``Who issued CD Special?''.

Below we provide examples of the prompts  used for the generation of \texttt{CREATE TABLE} and \texttt{INSERT INTO} SQL statements for different ambiguity types and configurations. We indicate placeholders we substitute with specific values in grey. For databases with attachment and vague ambiguities, we provide the prompts corresponding to the configurations shown in Figures~\ref{fig:dbs_attach}a and \ref{fig:dbs_vague}a, for the sake of  brevity.

Each configuration has strict conditions we can validate when executing predicted SQL statements. For instance, in the case of scope ambiguity, we automatically discard a database if there is no connection between the ``\texttt{Entity}'' and ``\texttt{Components}'' tables. However, we relaxed some restrictions to generate more efficiently, as they can be satisfied by automatic modifications later. For example, we require a value in ``\texttt{Components}'' to be connected to multiple, but not necessarily all, values in ``\texttt{Entity},'' as we can later add the missing connections.
\newpage
\begin{multicols}{2}
    \begin{tcolorbox}[colback=white, colframe=ivory, left=2pt,  coltitle=black, title=\textbf{Scope}: \texttt{CREATE TABLE}, halign=flush left]
        \footnotesize
    Create multiple connected tables in the \textcolor{gray}{\texttt{DOMAIN}} domain via SQLite, including a "\textcolor{gray}{\texttt{Entity}}" table, a "\textcolor{gray}{\texttt{Components}}" table (values: "\textcolor{gray}{\texttt{Specific\_Component}}" and others), and a "\textcolor{gray}{\texttt{Entities\_Components}}" table that joins these two tables (multiple rows of "\textcolor{gray}{\texttt{Entity}}" are connected to "\textcolor{gray}{\texttt{Specific\_Component}}" and other values of "\textcolor{gray}{\texttt{Components}}"). \\\vspace{3mm}
Other tables and columns are arbitrary. Each table must contain at least 3 columns.\\\vspace{3mm}
Provide multiple \texttt{CREATE TABLE} statements in SQLite.
    \end{tcolorbox}

    \begin{tcolorbox}[colback=white, colframe=ivory, coltitle=black, title=\textbf{Scope}: \texttt{INSERT INTO}, halign=flush left]
        \footnotesize
    SQLite database in the \textcolor{gray}{\texttt{DOMAIN}}  domain:\\\vspace{3mm}
\textcolor{gray}{\texttt{CREATE TABLE statements}}\\\vspace{3mm}
Insert 5 or more rows into each table.\\\vspace{3mm}
"\textcolor{gray}{\texttt{Components}}" table must contain "\textcolor{gray}{\texttt{Specific\_Component}}" and other values. All rows of table "\textcolor{gray}{\texttt{Entity}}" must be linked to two rows of "\textcolor{gray}{\texttt{Components}}", with "\textcolor{gray}{\texttt{Specific\_Component}}" and with another value, through "\textcolor{gray}{\texttt{Entities\_Components}}".\\\vspace{3mm}
Provide 5 or more \texttt{INSERT INTO} statements for each table in SQLite.
    \end{tcolorbox}
\end{multicols}
\vspace{-8mm}
\begin{multicols}{2}
    \begin{tcolorbox}[colback=white, colframe=ivory, left=2pt,  coltitle=black, title={\textbf{Attachment (a.)}: \texttt{CREATE TABLE}}, breakable, halign=flush left]
        \footnotesize
    Create multiple connected tables in the \textcolor{gray}{\texttt{DOMAIN}} domain via SQLite.  One table contains information about "\textcolor{gray}{\texttt{Class\_1}}" and "\textcolor{gray}{\texttt{Common\_Property}}" (containing values such as "\textcolor{gray}{\texttt{Common\_Value}}", etc.), and the other table contains information about "\textcolor{gray}{\texttt{Class\_2}}" and "\textcolor{gray}{\texttt{Common\_Property}}".\\\vspace{3mm}
Other tables and columns are arbitrary. Each table must contain at least 3 columns.\\\vspace{3mm}
Provide multiple \texttt{CREATE TABLE} statements in SQLite.
    \end{tcolorbox}

    \begin{tcolorbox}[colback=white, colframe=ivory, coltitle=black, title=\textbf{Attachment (a.)}: \texttt{INSERT INTO}, breakable, halign=flush left]
        \footnotesize
SQLite database in the \textcolor{gray}{\texttt{DOMAIN}}  domain:\\\vspace{3mm}
\text{\textcolor{gray}{\texttt{CREATE TABLE statements}}}\\\vspace{3mm}
Insert 5 or more rows into each table.\\\vspace{3mm}
There must be:
\begin{itemize}[leftmargin=3mm]  
\item  One row with "\textcolor{gray}{\texttt{Class\_1}}" and "\textcolor{gray}{\texttt{Common\_Property}}" equal to "\textcolor{gray}{\texttt{Common\_Value}}",
\item One row with "\textcolor{gray}{\texttt{Class\_1}}" and "\textcolor{gray}{\texttt{Common\_Property}}" different from "\textcolor{gray}{\texttt{Common\_Value}}",
\item  One row with "\textcolor{gray}{\texttt{Class\_2}}" and "\textcolor{gray}{\texttt{Common\_Property}}" equal to "\textcolor{gray}{\texttt{Common\_Value}}",
\item One row with "\textcolor{gray}{\texttt{Class\_2}}" and "\textcolor{gray}{\texttt{Common\_Property}}" different from "\textcolor{gray}{\texttt{Common\_Value}}",
\end{itemize}
Other rows are arbitrary. Use non-trivial values.\\\vspace{3mm}
Provide 5 or more \texttt{INSERT INTO} statements for each table in SQLite.
\end{tcolorbox}
\end{multicols}
\vspace{-8mm}
\begin{multicols}{2}
    \begin{tcolorbox}[colback=white, colframe=ivory, left=2pt,  coltitle=black, title={\textbf{Vague (a.)}: \texttt{CREATE TABLE}}, breakable, halign=flush left]
        \footnotesize
       Create multiple connected tables in the \textcolor{gray}{\texttt{DOMAIN}} domain via SQLite, including a table "\textcolor{gray}{\texttt{General\_Category\_1}}", a table "\textcolor{gray}{\texttt{General\_Category\_2}}" and a table "\textcolor{gray}{\texttt{Common\_Subject}}" connected to the tables "\textcolor{gray}{\texttt{General\_Category\_1}}" and "\textcolor{gray}{\texttt{General\_Category\_2}}". \\\vspace{3mm}
Other tables and columns are arbitrary. Each table must contain at least 3 columns.\\\vspace{3mm}
Provide multiple \texttt{CREATE TABLE} statements in SQLite.
    \end{tcolorbox}

    \begin{tcolorbox}[colback=white, colframe=ivory, coltitle=black, title=\textbf{Vague (a.)}: \texttt{INSERT INTO}, breakable, halign=flush left]
        \footnotesize
SQLite database in the \textcolor{gray}{\texttt{DOMAIN}}  domain:\\\vspace{3mm}
\text{\textcolor{gray}{\texttt{CREATE TABLE statements}}}\\\vspace{3mm}
Insert 5 or more rows into each table.\\\vspace{3mm}
Provide 5 or more \texttt{INSERT INTO} statements for each table in SQLite.
\end{tcolorbox}
\end{multicols}

\section{Human Annotation\label{apdx:annotators}}
 We recruited Prolific annotators using the following criteria: English as the first and primary language, no language-related disorders or literacy difficulties,  knowledge of SQL and database management, a Prolific approval rate of 100\%, and residency in the UK, USA, Ireland, Australia, New Zealand, or Canada. We also created trial tasks to choose the best annotators. 
 Each annotator was paid £9 per hour. The total budget for annotation was £1,600. All submissions in Prolific are anonymous. All annotators participated voluntarily and had the option to return the task at any time if they chose not to complete it. We only collected  data that was specifically requested from the annotators.

The first group of annotators (20 in total), participated in text-only tasks. These tasks included writing questions with scope and attachment ambiguity and  unambiguous interpretations for all questions, including vague ones. We assigned 3--5 instances per person, which took 10--20 minutes to annotate. 
The second group (10 annotators in total), wrote SQL queries and corresponding questions that were later rendered vague. Each annotator completed 3 instances, taking an average of 20 minutes per instance.

\begin{figure}[t]
\includegraphics[width=\textwidth]{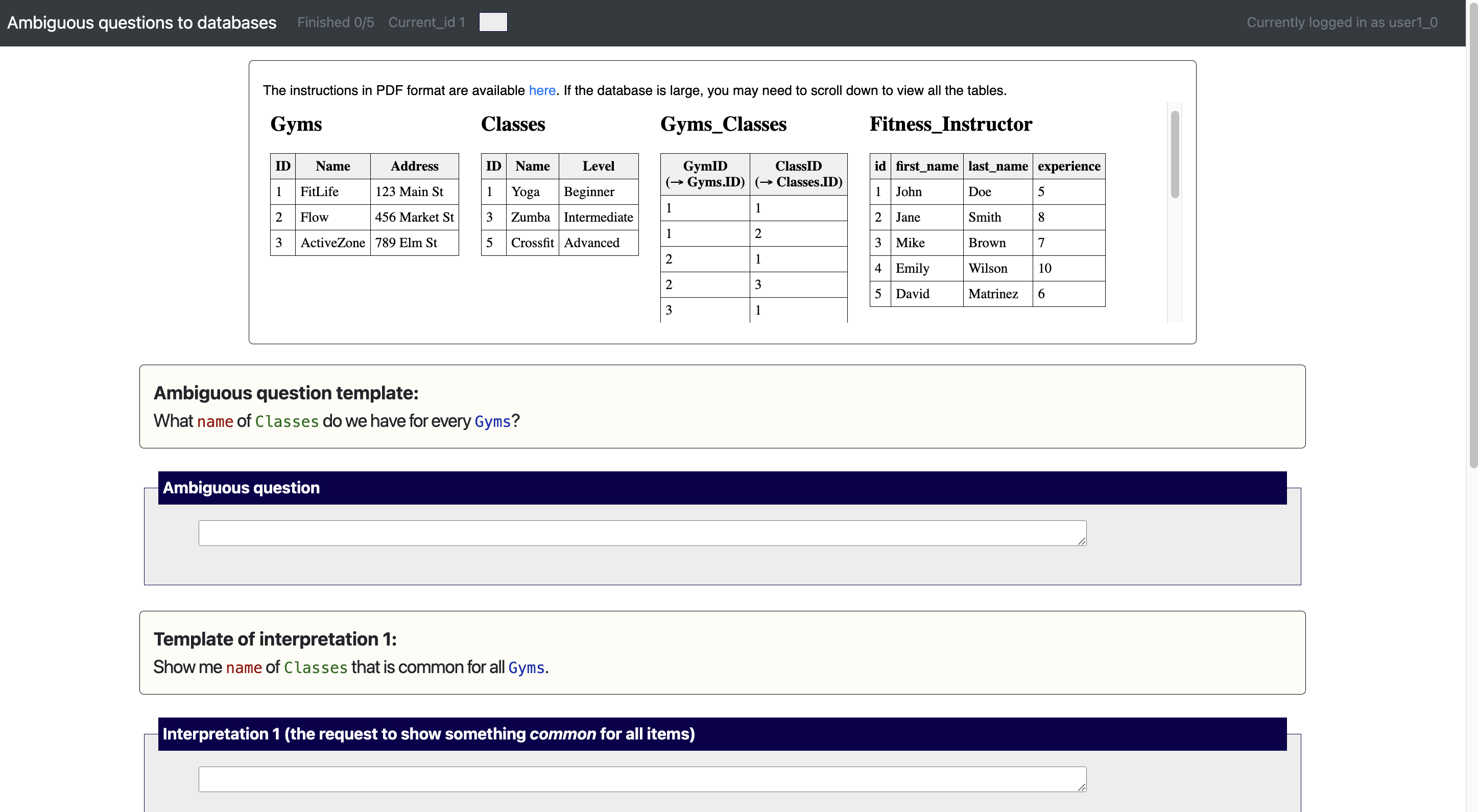}
\caption{Annotation Interface: example of writing questions with scope ambiguity and their unambiguous interpretations.}
\label{fig:interface}
\end{figure}

We provided separate instructions with examples for each task and manually reviewed all submissions to ensure all annotators followed them. We elicited annotations using the Potato tool \citep{pei-etal-2022-potato}.   An example of the interface is shown in Figure~\ref{fig:interface}. The code and detailed instructions used for annotation are available at: \url{https://github.com/saparina/ambrosia}.

\section{Implementation Details\label{apdx:implem_details}}
\paragraph{Models} We implemented the Prompt method using the Text Generation Inference toolkit\footnote{\url{https://github.com/huggingface/text-generation-inference}} for the Llama-family models; for OpenChat \citep{Wang-2024-openchat}, we used  the official implementation provided by the authors and the OpenAI API for the GPT models\footnote{\url{https://platform.openai.com/}}. We used the following versions of these models: openchat-3.5-1210, gpt-3.5-turbo-0125, gpt-4o-2024-05-13. The temperature for GPT models is 0. We implemented the Beam method using the vLLM library \citep{kwon2023efficient}. We used 4-8 CPUs, one NVIDIA-A100  GPU for OpenChat 7B and Llama3-8B and two NVIDIA-A100 GPUs for CodeLlama-70B and Llama3-70B. For the largest models, one zero-shot experiment (with fixed random seed) took one hour and one few-shot experiment with 5 examples took 5 hours.

\paragraph{Evaluation} To compute evaluation metrics on model predictions, we represent the output of an SQL query as a set of values. This allows matching queries that correspond to the same interpretation and yield the same data even if they are structured differently, e.g., with different column orders or different operations. 
For instance, some gold queries in the attachment category use the \texttt{UNION} operator, but they could also be written using the JOIN operator. Although this relaxation might result in occasional false positives, we found it rarely happens  as different SQL queries produce completely different outputs on our databases. We filter exact duplicates in predictions.

We measure precision as the number of correct queries in the output divided by the total number of predictions. Below we provide some examples:

Ambiguous case:
\begin{itemize}
    \item   If the model predicts only one SQL query from the gold set (one interpretation), precision is 1.0.
    \item If the model predicts all SQL queries from the gold set (all interpretations), precision remains 1.0.
    \item If multiple predictions include incorrect queries, precision falls between 0 and 1 (e.g., 0.5 might indicate one correct interpretation and one incorrect query, or two correct interpretations and two incorrect queries).
\end{itemize}

Unambiguous case:
\begin{itemize}
\item We have one gold SQL query. If the model predicts only this query, precision is 1.0.
\item If the model predicts the correct query along with additional incorrect ones, precision is reduced (e.g., 1 / total number of predictions).
\end{itemize}
In both cases, if there are no correct queries, precision is 0. When there is only one correct prediction, precision is 1.

We measure recall as the number of correct queries in the output divided by the total number of ground truth queries (interpretations). Below we provide some examples:

Ambiguous case:
\begin{itemize}
    \item If the model predicts all SQL queries from the ground truth set (covering all interpretations), recall is 1.0, even if additional incorrect queries are present in the output.
    \item If any ground truth interpretation is missing in the predictions, recall is between 0 and 1. For example, if there are two gold interpretations and the model predicts only one, recall would be 0.5.
\end{itemize}

Unambiguous case:
\begin{itemize}
\item If the model predicts the single gold query correctly, recall is 1.0, regardless of additional incorrect predictions.
\end{itemize}
In both cases, if there are no correct queries, recall is 0.

\paragraph{Prompting}

Below we provide the prompts used for the Prompt and Beam methods. We follow best practices for designing text-to-SQL prompts for LLMs, such as adding instructions like ``Do not include any explanations, and do not select extra columns beyond those requested in the question''  to prevent models from generating redundant columns or descriptions \citep{dong2023c3, dail_sql}. However, we found that the models do not always adhere to these instructions. To address this, we implemented a separate extractor for SQL queries from each model's output. Additionally, we observed that models often generate separate SQL queries for attachment ambiguity instead of a single complex query with a \texttt{UNION} of two subqueries. To resolve this, we added the instruction ``Show the results in one table'' to each question in this category. 

\begin{multicols}{2}
    \begin{tcolorbox}[colback=white, colframe=green, left=2pt,  coltitle=black, title=\textbf{Prompt}, breakable, halign=flush left]
        \footnotesize
The task is to write SQL queries based on the provided questions in English. Questions can take the form of an instruction or command and can be ambiguous, meaning they can be interpreted in different ways. In such cases, write all possible SQL queries corresponding to different interpretations and separate each SQL query with an empty line. Do not include any explanations, and do not select extra columns beyond those requested in the question.\\\vspace{3mm}
Given the following SQLite database schema:\\\vspace{3mm}
\textcolor{gray}{\texttt{SQL\_DATABASE\_DUMP}}\\\vspace{3mm}
Answer the following:\\
\textcolor{gray}{\texttt{QUESTION}}
    \end{tcolorbox}

   \begin{tcolorbox}[colback=white, colframe=green, coltitle=black, title=\textbf{Beam}, breakable, halign=flush left]
        \footnotesize
The task is to write SQL queries based on the provided questions in English. Questions can take the form of an instruction or command. Do not include any explanations, and do not select extra columns beyond those requested in the question.\\\vspace{3mm}
Given the following SQLite database schema:\\\vspace{3mm}
\textcolor{gray}{\texttt{SQL\_DATABASE\_DUMP}}\\\vspace{3mm}
Answer the following: \\
\textcolor{gray}{\texttt{QUESTION}}
\end{tcolorbox}
\end{multicols}

Below we present the few-shot prompt with randomly selected
in-context examples. Each example includes ambiguous questions, their unambiguous interpretations,
and corresponding SQL queries:

\noindent\vspace*{0pt}\begin{tcolorbox}[colback=white, colframe=green, left=2pt,  coltitle=black, title=\textbf{Few-Shot Prompt}, breakable, halign=flush left]
    \footnotesize
The task is to write SQL queries based on the provided questions in English. Questions can take the form of an instruction or command and can be ambiguous, meaning they can be interpreted in different ways. In such cases, write all possible SQL queries corresponding to different interpretations and separate each SQL query with an empty line. Do not include any explanations, and do not select extra columns beyond those requested in the question.\\\vspace{3mm}
Some example databases, questions and corresponding SQL queries are provided based on similar problems:\\\vspace{3mm}
Example 1:\\\vspace{3mm}
Given the following SQLite database schema:\\\vspace{3mm}
\textcolor{nicecol}{\texttt{EXAMPLE\_SQL\_DATABASE\_DUMP}}\\\vspace{3mm}
Answer the following:\\
\textcolor{nicecol1}{\texttt{AMBIGUOUS\_QUESTION}}\\\vspace{3mm}
SQL query(s):\\
\textcolor{nicecol2}{\texttt{SQL\_QUERY\_1}}\\
\textcolor{nicecol3}{\texttt{SQL\_QUERY\_2}}\\
\textcolor{nicecol4}{\texttt{[OPTIONAL\_SQL\_QUERY\_3]}}\\\vspace{3mm}
Answer the following:\\
\textcolor{nicecol2}{\texttt{INTERPRETATION\_1}}\\\vspace{3mm}
SQL query(s):\\
\textcolor{nicecol2}{\texttt{SQL\_QUERY\_1}}\\\vspace{3mm}
Answer the following:\\
\textcolor{nicecol3}{\texttt{INTERPRETATION\_2}}\\\vspace{3mm}
SQL query(s):\\
\textcolor{nicecol3}{\texttt{SQL\_QUERY\_2}}\\
\dots\\\vspace{3mm}
Given the following SQLite database schema:\\\vspace{3mm}
\textcolor{gray}{\texttt{SQL\_DATABASE\_DUMP}}\\\vspace{3mm}
Answer the following:\\
\textcolor{gray}{\texttt{MAIN\_QUESTION}}
\end{tcolorbox}

Additionally, we consider the prompt with all three ambiguities and their definitions included:

\noindent\vspace*{0pt}\begin{tcolorbox}[colback=white, colframe=green, left=2pt,  coltitle=black, title=\textbf{Few-Shot Prompt with 3 Ambiguities and Definitions}, breakable, halign=flush left]
    \footnotesize
The task is to write SQL queries based on the provided questions in English. Questions can take the form of an instruction or command and can be ambiguous, meaning they can be interpreted in different ways. In such cases, write all possible SQL queries corresponding to different interpretations and separate each SQL query with an empty line. Do not include any explanations, and do not select extra columns beyond those requested in the question.\\\vspace{3mm}
Some example databases, questions and corresponding SQL queries are provided based on similar problems:\\\vspace{3mm}
Type 1: Scope Ambiguity\\\vspace{3mm}
Definition: Scope ambiguity issue arises when it is unclear which elements a quantifier, such as "each", "every" or "all", refers to.\\\vspace{3mm}
Example:\\\vspace{3mm}
Given the following SQLite database schema:\\\vspace{3mm}
\textcolor{nicecol}{\texttt{SCOPE\_SQL\_DATABASE\_DUMP}}\\\vspace{3mm}
Answer the following:\\
\textcolor{nicecol1}{\texttt{SCOPE\_QUESTION}}\\\vspace{3mm}
SQL query(s):\\
\textcolor{nicecol2}{\texttt{SQL\_QUERY\_1}}\\
\textcolor{nicecol3}{\texttt{SQL\_QUERY\_2}}\\\vspace{3mm}
Answer the following:\\
\textcolor{nicecol2}{\texttt{SCOPE\_INTERPRETATION\_1}}\\\vspace{3mm}
SQL query(s):\\
\textcolor{nicecol2}{\texttt{SQL\_QUERY\_1}}\\
\dots\\\vspace{3mm}
Type 2: Attachment Ambiguity\\\vspace{3mm}
Definition: Attachment ambiguity occurs when it is unclear how a modifier or phrase is attached to the rest of the sentence.\\\vspace{3mm}
\textcolor{nicecol4}{\texttt{ATTACHMENT\_EXAMPLE}}\\\vspace{3mm}
Type 3: Vagueness\\\vspace{3mm}
Definition: Vagueness occurs when context creates uncertainty about which set of entities is being referred to.\\\vspace{3mm}
\textcolor{nicecol4}{\texttt{VAGUE\_EXAMPLE}}\\\vspace{3mm}
Given the following SQLite database schema:\\\vspace{3mm}
\textcolor{gray}{\texttt{SQL\_DATABASE\_DUMP}}\\\vspace{3mm}
Answer the following:\\
\textcolor{gray}{\texttt{MAIN\_QUESTION}}
\end{tcolorbox}

\section{Additional Results \label{apdx:additional_results}}
\paragraph{Extended Analysis} Table~\ref{tab:num_predicted} presents the number of unique SQL queries predicted by the zero-shot Llama3-70B (Prompt) and the number of unique results obtained upon executing these predictions. As can be seen, the number of execution results is consistently lower than the number of predicted SQL queries, indicating that some queries, despite differing in structure, yield the same results. Typically, both metrics are around~2, implying that the model is capable of generating multiple distinct SQL queries that produce different results.
Although the figures for unambiguous questions are lower than for ambiguous ones, they still do not approach the optimum value of 1.0 (unambiguous questions correspond to a single SQL query).

To further explore how the model captures ambiguity, we consider the task of ambiguity detection. We present the model with a question and ask it to determine whether it  is ambiguous.
Similar to the Prompt method, we provide the model with comprehensive instructions regarding this task:
\begin{tcolorbox}[colback=white, colframe=green, coltitle=black, title=\textbf{Ambiguity Detection}, breakable, halign=flush left]
        \footnotesize
The task is to identify ambiguous questions in English that are intended to interact with an SQLite database. Questions can take the form of an instruction or command and can be ambiguous, meaning they can be interpreted in different ways (corresponding to different SQL queries that produce different results). Answer Yes or No, and do not include any explanations.\\\vspace{3mm}
Given the following SQLite database schema:\\\vspace{3mm}
\textcolor{gray}{\texttt{SQL\_DATABASE\_DUMP}}\\\vspace{3mm}
Is the following question ambiguous: \\
\textcolor{gray}{\texttt{QUESTION}}
\end{tcolorbox}

Table~\ref{tab:detect_ambig} shows the accuracy of  zero-shot Llama3-70B in detecting ambiguity. Our results indicate that the model tends to overestimate ambiguity, classifying questions as ambiguous in approximately 80\% of cases, regardless of whether they  genuinely are. This is a limitation of prompt-based approaches, which may confuse the model rather than provide helpful guidance.

\begin{table}[t]
    \begin{minipage}{.58\linewidth}
\caption{Number of unique SQL queries predicted by zero-shot Llama3-70B (Prompt) and number of unique results after execution of these queries. The model predicts more than one query even for unambiguous questions.}
\label{tab:num_predicted}
      \centering
\footnotesize
\begin{tabular}{cccc}
  \multicolumn{2}{c}{\bf \# Unique SQL Queries} & \multicolumn{2}{c}{\bf \# Unique Exec Results}  \\
         \textbf{ambig} & \textbf{unambig} & \textbf{ambig} &
  \textbf{unambig} \\  \toprule
  2.4& 2.3 & 2.0 & 1.7 \\ %
  \bottomrule
\end{tabular}
\end{minipage}%
\hfill
    \begin{minipage}{.38\linewidth}
    \caption{Zero-shot accuracy of the Llama3-70B in detecting ambiguity in questions. The model overestimates ambiguity.}
    \label{tab:detect_ambig}
    \centering
\footnotesize
\begin{tabular}{cc}
             \multicolumn{2}{c}{\bf \% Accuracy} \\
   \textbf{ambig} & \textbf{unambig} \\  \toprule
     81.2   & 26.1  \\ \bottomrule %
\end{tabular}
\end{minipage} 
\end{table}

\paragraph{Error Cases} 
We conducted an error analysis on the zero-shot Llama3-70B (Prompt) by randomly selecting 50 examples where the model failed. Our analysis revealed errors typical of standard text-to-SQL parsing, as well as errors specifically related to ambiguity. Common text-to-SQL errors included: incorrect output columns (29\%), including superfluous columns contrary to our instructions, errors in the filters and joins (17\%), failure to include a UNION of two queries (17\%, typical for the attachment category), and execution errors (5\%). The most frequent error relating to ambiguity is the generation of different variations of the same SQL query, which occurred in 34\% of cases, even for unambiguous questions. For ambiguous queries, the model provided only one interpretation in 11\% of cases. See the examples provided next.

\begin{table*}[h]
\footnotesize
\centering
\settowidth\tymin{\textbf{Predictions}}
\begin{tabulary}{\textwidth}{@{}L@{\;\;}L@{}}
\toprule
\multicolumn{2}{c}{\textbf{Type:} Scope  \qquad \textbf{Domain:} Agriculture \qquad \textbf{Ambiguous?} Yes}  \\\addlinespace[5pt] %
\textbf{Question} & What kind of crop does each farm offer?\\\addlinespace[5pt]
\textbf{Predictions}
& \cmark \texttt{ SELECT \textcolor{nicegreen}{f.Name, c.CropName} FROM Farms f JOIN FarmsCrops fc ON f.FarmID = fc.FarmID JOIN Crops c ON fc.CropID = c.CropID}\\\addlinespace[3pt]
& \cmark \texttt{ SELECT \textcolor{nicegreen}{DISTINCT f.Name, c.CropName} FROM Farms f JOIN FarmsCrops fc ON f.FarmID = fc.FarmID JOIN Crops c ON fc.CropID = c.CropID} \\\addlinespace[5pt]
\textbf{Gold} & 1. \texttt{SELECT \textcolor{niceblue}{c.CropName} FROM FarmsCrops fc JOIN Crops c ON fc.CropID = c.CropID JOIN Farms f ON fc.FarmID = f.FranID \textcolor{niceblue}{GROUP BY fc.CropID HAVING COUNT(DISTINCT fc.FarmID) = (SELECT COUNT(*) FROM Farms)}
}\\\addlinespace[3pt]
& 2. \texttt{SELECT \textcolor{niceblue}{f.name, c.cropName} FROM FarmsCrops fc JOIN Crops c ON fc.CropID = c.CropID JOIN Farms f ON fc.FarmID = f.FarmID
}\\\addlinespace[5pt]
\textbf{Comments} & All predicted queries are correct but they all provide the same information and correspond to one interpretation (distributive; gold query 2). \\\midrule
\multicolumn{2}{c}{\textbf{Type:} Scope  \qquad \textbf{Domain:} Healthcare \qquad \textbf{Ambiguous?} No}  \\\addlinespace[5pt] %
\textbf{Question} & For every program, list the activity descriptions\\\addlinespace[5pt]
\textbf{Predictions}
& \xmark \texttt{ SELECT \textcolor{nicered}{hpp.description, a.description} FROM Health\_Promotion\_Program hpp JOIN Health\_Promotion\_Program\_Activities hppa ON hpp.program\_id = hppa.program\_id JOIN Activities a ON hppa.activity\_id = a.activity\_id}\\\addlinespace[3pt]
& \xmark \texttt{ SELECT \textcolor{nicered}{a.description} FROM Health\_Promotion\_Program hpp JOIN Health\_Promotion\_Program\_Activities hppa ON hpp.program\_id = hppa.program\_id JOIN Activities a ON hppa.activity\_id = a.activity\_id GROUP BY hpp.program\_id} \\\addlinespace[5pt]
\textbf{Gold} & \texttt{SELECT \textcolor{nicegreen}{hpp.name, a.description} FROM Health\_Promotion\_Program\_Activities hppa JOIN Activities a ON hppa.activity\_id = a.activity\_id JOIN Health\_Promotion\_Program hpp ON hppa.program\_id = hpp.program\_id
}\\\addlinespace[5pt]
\textbf{Comments} & The first predicted query has an error in the output columns: it outputs the program description instead of the program name. The second query omits the program name column altogether.\\\midrule
\multicolumn{2}{c}{\textbf{Type:} Attachment  \qquad \textbf{Domain:} Filmmaking \qquad \textbf{Ambiguous?} Yes}  \\\addlinespace[5pt]
\textbf{Question} & Provide all sound designs and music scores created by Thomas Newman. Show them in one table.\\\addlinespace[5pt]
\textbf{Predictions} & \xmark \texttt{ SELECT description FROM \textcolor{nicered}{SoundDesigns} WHERE creatorId = (SELECT id FROM Creators WHERE name = \textcolor{nicered}{'Thomas Newman'})}\\\addlinespace[3pt]
& \xmark \texttt{ SELECT description FROM \textcolor{nicered}{MusicScores} WHERE creatorId = (SELECT id FROM Creators WHERE name = \textcolor{nicered}{'Thomas Newman'})} \\\addlinespace[3pt]
& \cmark \texttt{ SELECT sd.description, ms.description FROM \textcolor{nicegreen}{SoundDesigns sd, MusicScores ms} WHERE sd.creatorId = ms.creatorId AND sd.creatorId = (SELECT id FROM Creators WHERE name = \textcolor{nicegreen}{'Thomas Newman'})}\\\addlinespace[3pt]
& \cmark \texttt{ SELECT * FROM (SELECT description FROM \textcolor{nicegreen}{SoundDesigns} WHERE creatorId = (SELECT id FROM Creators WHERE name = \textcolor{nicegreen}{'Thomas Newman'})) UNION SELECT * FROM (SELECT description FROM \textcolor{nicegreen}{MusicScores} WHERE creatorId = (SELECT id FROM Creators WHERE name = \textcolor{nicegreen}{'Thomas Newman'}))}\\\addlinespace[5pt]
\textbf{Gold} & 1. \texttt{SELECT sd.description FROM Creators c JOIN \textcolor{niceblue}{SoundDesigns} sd ON c.id = sd.creatorId WHERE c.name = \textcolor{niceblue}{"Thomas Newman"} UNION SELECT ms.description FROM Creators c JOIN \textcolor{niceblue}{MusicScores} ms ON c.id = ms.creatorInterId WHERE c.name = \textcolor{niceblue}{"Thomas Newman"}} \\\addlinespace[3pt]
& 2. \texttt{SELECT sd.description FROM Creators c JOIN \textcolor{niceblue}{SoundDesigns} sd ON c.id = sd.creatorId UNION SELECT ms.description FROM Creators c JOIN \textcolor{niceblue}{MusicScores} ms ON c.id = ms.creatorId WHERE c.name = \textcolor{niceblue}{"Thomas Newman"}} \\\addlinespace[5pt]
\textbf{Comments} & The first two predictions should be subqueries of a single complex query (failure to include a UNION of two queries). The last two queries are correct; however, they provide the same information and correspond to one interpretation (with high attachment; gold query 1). %
 \\\bottomrule
 \end{tabulary}
 \end{table*}
\begin{table*}[h]
\footnotesize
\centering
\settowidth\tymin{\textbf{Predictions}}
\begin{tabulary}{\textwidth}{@{}L@{\;\;}L@{}}
\toprule
\multicolumn{2}{c}{\textbf{Type:} Attachment  \qquad \textbf{Domain:} Hospitality \qquad \textbf{Ambiguous?} No}  \\\addlinespace[5pt] %
\textbf{Question} & Show all youth hostels and also show the hotels, but only those with 50-priced rooms. Show them in one table.\\\addlinespace[5pt]
\textbf{Predictions}
& \xmark \texttt{ SELECT \textcolor{nicered}{name, location, accommodation\_type}
FROM hotels
WHERE accommodation\_type = \textcolor{nicered}{'Youth Hostel'}}\\\addlinespace[3pt]
& \xmark \texttt{ SELECT \textcolor{nicered}{name, location, accommodation\_type} FROM hotels WHERE id IN (SELECT hotel\_id  FROM rooms WHERE price = \textcolor{nicered}{50.0})} \\\addlinespace[3pt]
& \xmark \texttt{ SELECT \textcolor{nicered}{name, location, accommodation\_type}
FROM hotels
WHERE accommodation\_type = \textcolor{nicegreen}{'Youth Hostel'}
UNION
SELECT \textcolor{nicered}{name, location, accommodation\_type}
FROM hotels
WHERE id IN (
  SELECT hotel\_id
  FROM rooms
  WHERE price = \textcolor{nicered}{50.0}
)}\\\addlinespace[5pt]
\textbf{Gold} & \texttt{SELECT \textcolor{niceblue}{h.name} FROM hotels h JOIN rooms r ON h.id = r.hotel\_id WHERE h.accommodation\_type = \textcolor{niceblue}{"Youth Hostel" OR} (h.accommodation\_type = \textcolor{niceblue}{"Hotel" AND} r.price = \textcolor{niceblue}{50.0})
}\\\addlinespace[5pt]
\textbf{Comments} & All predicted queries have redundant output columns. The first two predictions should be subqueries of a single complex query (failure to include a UNION of two queries).  \\\midrule
\multicolumn{2}{c}{\textbf{Type:} Vague  \qquad \textbf{Domain:} Journalism \qquad \textbf{Ambiguous?} Yes}  \\\addlinespace[5pt] %
\textbf{Question} & List all the articles with their respective editor and author names.\\\addlinespace[5pt]
\textbf{Predictions}
& \cmark \texttt{ SELECT a.title, \textcolor{nicegreen}{me.name AS editor}, au.fullname AS author
FROM Articles a
JOIN Final\_Draft fd ON a.final\_draft\_id = fd.id
JOIN \textcolor{nicegreen}{Managing\_Editor me} ON fd.managing\_editor\_id = me.id
JOIN Authors au ON a.author\_id = au.author\_id}\\\addlinespace[3pt]
& \cmark \texttt{ SELECT a.title, \textcolor{nicegreen}{ce.name AS editor}, au.fullname AS author
FROM Articles a
JOIN Final\_Draft fd ON a.final\_draft\_id = fd.id
JOIN \textcolor{nicegreen}{Copy\_Editor ce} ON fd.copy\_editor\_id = ce.id
JOIN Authors au ON a.author\_id = au.author\_id} \\\addlinespace[5pt]
\textbf{Gold} & 1. \texttt{SELECT A.title, \textcolor{niceblue}{ME.name}, AU.fullname FROM Articles A JOIN Final\_Draft FD ON FD.id = A.final\_draft\_id JOIN \textcolor{niceblue}{Managing\_Editor ME} ON ME.id = FD.managing\_editor\_id JOIN Authors AU ON AU.author\_id = A.author\_id
}\\\addlinespace[3pt]
& 2. \texttt{SELECT A.title, \textcolor{niceblue}{CE.name}, AU.fullname FROM Articles A JOIN Final\_Draft FD ON FD.id = A.final\_draft\_id JOIN \textcolor{niceblue}{Copy\_Editor CE} ON CE.id = FD.Copy\_Editor\_id JOIN Authors AU ON AU.author\_id = A.author\_id
}\\\addlinespace[3pt]
 & 3. \texttt{SELECT A.title, \textcolor{niceblue}{ME.name, CE.name}, AU.fullname FROM Articles A JOIN Final\_Draft FD ON FD.id = A.final\_draft\_id JOIN \textcolor{niceblue}{Managing\_Editor ME} ON ME.id = FD.managing\_editor\_id JOIN Authors AU ON AU.author\_id = A.author\_id JOIN \textcolor{niceblue}{Copy\_Editor CE} ON CE.id = FD.Copy\_Editor\_id
 }\\\addlinespace[5pt]
\textbf{Comments} & All predicted queries are correct and correspond to two different interpretations (Component), but the third interpretation (Full) is missing.  \\\midrule
\multicolumn{2}{c}{\textbf{Type:} Vague  \qquad \textbf{Domain:} Student \qquad \textbf{Ambiguous?} No}  \\\addlinespace[5pt] %
\textbf{Question} & List students born after 1990, along with their field of study.\\\addlinespace[5pt]
\textbf{Predictions}
& \xmark \texttt{ SELECT fullname, Field\_Of\_Study FROM Students WHERE \textcolor{nicered}{BirthDate > '1990-01-01'}}\\\addlinespace[3pt]
& \xmark \texttt{ SELECT fullname, Field\_Of\_Study FROM Students WHERE \textcolor{nicered}{YEAR(BirthDate) > 1990}} \\\addlinespace[5pt]
\textbf{Gold} & \texttt{SELECT fullname, Field\_Of\_Study FROM Students WHERE \textcolor{niceblue}{CAST(SUBSTR(BirthDate, 1, 4) AS INT)  > '1990'}}\\\addlinespace[5pt]
\textbf{Comments} & The first predicted query has an error in its filter as it incorrectly includes students born in 1990. The second query is not executable as SQLite does not have a built-in \texttt{YEAR} function.  \\\bottomrule
\end{tabulary}
\end{table*}

\cleardoublepage
\section{\texorpdfstring{Datasheet for $\varmathbb{AMBROSIA}$}{~}\label{apdx:datasheet}}
Below we provide a datasheet \citep{datasheets} for \dataset{}.

\subsection*{Motivation}
\paragraph{For what purpose was the dataset created?}
\dataset{} aims to study ambiguity in semantic parsing, with a focus on text-to-SQL parsing due to its real-world relevance. Large Language Models (LLMs) have demonstrated strong performance on existing unambiguous text-to-SQL benchmarks, allowing us to concentrate specifically on studying ambiguity. Unlike other logical forms, such as the first-order logic used by \citet{eskin-2024-zero}, SQL queries are straightforward to execute and verify, making them ideal for testing. We aim to replicate practical semantic parsing scenarios with diverse databases in various domains that support ambiguity in questions. We also cover three different types of ambiguity and provide human-written interpretations that enable in-depth analysis. This distinguishes \dataset{} from other benchmarks \citep{wang-etal-2023-know, bhaskar-etal-2023-benchmarking}, which rely on augmentations of existing datasets, and are thus less realistic, and lack diversity.

\paragraph{Who created the dataset (e.g., which team, research group) and on behalf of which entity (e.g., company, institution, organization)?} It was created by the authors.

\paragraph{Who funded the creation of the dataset?} We gratefully acknowledge  the support of the UK Engineering and Physical Sciences Research Council
(grant EP/W002876/1).

\subsection*{Composition}
\paragraph{What do the instances that comprise the dataset represent (e.g., documents, photos, people, countries)?} \dataset{} includes databases, ambiguous questions and requests related to them, their corresponding interpretations and SQL queries.

\paragraph{How many instances are there in total (of each type, if appropriate)?}  \dataset{}  contains
846~multi-table databases in 16~distinct domains, 1,277 ambiguous questions,  their unambiguous interpretations provided by humans 
and complex SQL queries (2,965 in total).

\paragraph{Does the dataset contain all possible instances or is it a sample (not necessarily random) of instances from a larger set?} 
The dataset includes all instances we collected.  Our data collection approach allows for future expansion of the dataset.

\paragraph{What data does each instance consist of?}
One example in \dataset{} includes a database that supports ambiguity, an ambiguous question, its possible interpretations, and SQL queries corresponding to these interpretations. Interpretations can be viewed separately as unambiguous questions.

\paragraph{Is there a label or target associated with each instance?}
For each question, the targets are the corresponding SQL queries: one query for an unambiguous question and 2--3 queries for an ambiguous one. Additionally, each database is labeled based on the domain it represents, its configuration, and the type of ambiguity present in the relevant questions.

\paragraph{Is any information missing from individual instances?} No

\paragraph{Are relationships between individual instances made explicit (e.g., users’ movie ratings, social network links)?}
Annotators were given databases for context when writing questions, interpretations, and SQL queries for vague questions. Other SQL queries were constructed to query the given database. Annotators provided interpretations when they were aware of the ambiguous question they related to.

\paragraph{Are there recommended data splits (e.g., training, development/validation, testing)?}
We reserve 10\% for a few-shot learning scenario and recommend to evaluate on the remaining data.

\paragraph{Are there any errors, sources of noise, or redundancies in the dataset?}
We manually validate all databases and human annotations. However, there might be annotation errors that we did not notice, such as interpretations that fail to disambiguate the questions or questions that are not ambiguous. The generated database might also contain tables that are irrelevant to the chosen key concepts.

\paragraph{Is the dataset self-contained, or does it link to or otherwise rely on external resources (e.g., websites, tweets, other datasets)?} The dataset is self-contained.

\paragraph{Does the dataset contain data that might be considered confidential (e.g., data that is protected by legal privilege or by doctor-patient confidentiality, data that includes the content of individuals’ non-public communications)?}
No.

\paragraph{Does the dataset contain data that, if viewed directly, might be offensive, insulting, threatening, or might otherwise cause anxiety?}
No.

\paragraph{Does the dataset relate to people?}
Yes.

\paragraph{Does the dataset identify any subpopulations (e.g., by age, gender)?}
There might be questions or databases containing information about age or gender, but they do not provide any opinions.

\paragraph{Is it possible to identify individuals (i.e., one or more natural persons), either directly or indirectly (i.e., in combination with other data) from the dataset?}
No, all annotation submissions were fully anonymous.

\paragraph{Does the dataset contain data that might be considered sensitive in any way (e.g., data that reveals racial or ethnic origins, sexual orientations, religious beliefs, political opinions or \texttt{UNION} memberships, or locations; financial or health data; biometric or genetic data; forms of government identification, such as social security numbers; criminal history)?}
We manually verify that the dataset does not contain any sensitive or harmful information.

\subsection*{Collection Process}

\paragraph{How was the data associated with each instance acquired?}

 Human annotators wrote ambiguous questions along with their unambiguous interpretations. Databases were generated by the LLM OpenChat (\citep{Wang-2024-openchat}), and SQL queries were automatically created using templates to address scope and attachment ambiguity. Annotators were asked to write SQL queries and corresponding questions for databases, which we had modified by merging vague concepts into more general ones, thereby removing ambiguity. After the annotation was complete, we restored the original databases and adjusted the SQL queries. The questions became vague due to the reintroduction of vague concepts. We manually verified questions, their interpretations and databases. SQL queries were executed to validate that they produced non-empty, distinct results. See details in Section 3 and Appendix E.

\paragraph{What mechanisms or procedures were used to collect the data (e.g., hardware apparatus or sensor, manual human curation, software program, software API)?}
We elicited annotations using the Potato tool \citep{pei-etal-2022-potato}.

\paragraph{If the dataset is a sample from a larger set, what was the sampling strategy (e.g., deterministic, probabilistic with specific sampling probabilities)?} N/A

\paragraph{Who was involved in the data collection process (e.g., students, crowdworkers, contractors) and how were they compensated (e.g., how much were crowdworkers paid)?} We recruited annotators through the Prolific crowdsourcing platform. Each annotator was paid £9 per hour.

\paragraph{Over what timeframe was the data collected?}
October 2023--April 2024.

\paragraph{Were any ethical review processes conducted (e.g., by an institutional review board)?}
No.

\paragraph{Does the dataset relate to people?}
Yes.

\paragraph{Did you collect the data from the individuals in question directly, or obtain it via third parties or other sources (e.g., websites)?} The data was collected directly.

\paragraph{Were the individuals in question notified about the data collection?}
Yes, by requirement process in Prolific.

\paragraph{Did the individuals in question consent to the collection and use of their data?}
Yes, by requirement process in Prolific.

\paragraph{If consent was obtained, were the consenting individuals provided with a mechanism to revoke their consent in the future or for certain uses?}
No.

\paragraph{Has an analysis of the potential impact of the dataset and its use on data subjects (e.g., a data protection impact analysis) been conducted?}
No, all submissions on Prolific are anonymous, and we did not collect any data beyond what was requested.

\subsection*{Preprocessing/Cleaning/Labeling}

\paragraph{Was any preprocessing/cleaning/labeling of the data done (e.g., discretization or bucketing, tokenization, part-of-speech tagging, SIFT feature extraction, removal of instances, processing of missing values)?}
We manually validated all generated concepts and databases and human-written submissions. 

\paragraph{Was the “raw” data saved in addition to the preprocessed/cleaned/labeled data (e.g., to support unanticipated future uses)?}
No.

\paragraph{Is the software used to preprocess/clean/label the instances available?}
Yes, see \url{https://github.com/saparina/ambrosia}

\subsection*{Uses}

\paragraph{Has the dataset been used for any tasks already?}
Yes, we benchmarked LLMs on the text-to-SQL semantic parsing with ambiguous questions.

\paragraph{Is there a repository that links to any or all papers or systems that use the dataset?}
No.

\paragraph{What (other) tasks could the dataset be used for?}
It can be used for studying different types of generalization, e.g.,~across domains or types of questions.

\paragraph{Is there anything about the composition of the dataset or the way it was collected and preprocessed/cleaned/labeled that might impact future uses?}
We cannot guarantee that the dataset is free from errors (see Section~5 for limitations). Additionally, the recruitment of annotators with knowledge of database management and SQL introduces a potential bias. Future research could mitigate this by paraphrasing the questions.

\paragraph{Are there tasks for which the dataset should not be used?}
No.

\subsection*{Distribution}
\paragraph{Will the dataset be distributed to third parties outside of the entity (e.g., company, institution, organization) on behalf of which the dataset was created?}
Yes, the dataset is publicly available.

\paragraph{How will the dataset be distributed (e.g., tarball on website, API, GitHub)?}
The dataset is available at  \url{ambrosia-benchmark.github.io}

\paragraph{When will the dataset be distributed?}
Now.

\paragraph{Will the dataset be distributed under a copyright or other intellectual property (IP) license, and/or under applicable terms of use (ToU)?}
The dataset is distributed under CC BY 4.0

\paragraph{Have any third parties imposed IP-based or other restrictions on the data associated with the instances?}
No.

\paragraph{Do any export controls or other regulatory restrictions apply to the dataset or to individual instances?}
No.

\subsection*{Maintenance}
\paragraph{Who is supporting/hosting/maintaining the dataset?} The authors of this paper.

\paragraph{How can the owner/curator/manager of the dataset be contacted (e.g., email address)?} The first author of this paper.

\paragraph{Is there an erratum?}
No.

\paragraph{Will the dataset be updated (e.g., to correct labeling errors, add new instances, delete instances)?}
We plan to update the dataset by correcting any annotation errors as they are identified. The new version will be labeled accordingly.

\paragraph{If the dataset relates to people, are there applicable limits on the retention of the data associated with the instances (e.g., were individuals in question told that their data would be retained for a fixed period of time and then deleted)?}
No.

\paragraph{Will older versions of the dataset continue to be supported/hosted/maintained?}
No.

\paragraph{If others want to extend/augment/build on/contribute to the dataset, is there a mechanism for them to do so?}
Yes, the code used for data collection is available at \url{https://github.com/saparina/ambrosia}.

\end{document}